\documentclass{article}

\usepackage{microtype}
\usepackage{graphicx}
\usepackage{subfigure}
\usepackage{caption}
\usepackage{booktabs} 
\usepackage{adjustbox}
\usepackage{subcaption}
\usepackage{bbm}
\usepackage{float}
\usepackage{pifont}
\usepackage{multirow}
\usepackage{hyperref}

\usepackage[accepted]{icml2024}

\usepackage{amsmath}
\usepackage{amssymb}
\usepackage{mathtools}
\usepackage{amsthm}
\usepackage{amsmath,amssymb}

\newcommand{\cmark}{\ding{51}}%
\newcommand{\xmark}{\ding{55}}%
\usepackage{tikz}

\makeatletter

\usetikzlibrary{shapes}
\usetikzlibrary{fit}
\usetikzlibrary{chains}
\usetikzlibrary{arrows}

\tikzstyle{latent} = [circle,fill=white,draw=black,inner sep=1pt,
minimum size=20pt, font=\fontsize{10}{10}\selectfont, node distance=1]
\tikzstyle{obs} = [latent,fill=gray!25]
\tikzstyle{const} = [rectangle, inner sep=0pt, node distance=1]
\tikzstyle{factor} = [rectangle, fill=black,minimum size=5pt, inner
sep=0pt, node distance=0.4]
\tikzstyle{det} = [latent, diamond]

\tikzstyle{plate} = [draw, rectangle, rounded corners, fit=#1]
\tikzstyle{wrap} = [inner sep=0pt, fit=#1]
\tikzstyle{gate} = [draw, rectangle, dashed, fit=#1]

\tikzstyle{caption} = [font=\footnotesize, node distance=0] %
\tikzstyle{plate caption} = [caption, node distance=0, inner sep=0pt,
below left=5pt and 0pt of #1.south east] %
\tikzstyle{factor caption} = [caption] %
\tikzstyle{every label} += [caption] %

\newcommand{\edge}[3][]{ %
  \foreach \x in {#2} { %
    \foreach \y in {#3} { %
      \path (\x) edge [->, >={triangle 45}, #1] (\y) ;%
    } ;
  } ;
}

\newcommand{\plate}[4][]{ %
  \node[wrap=#3] (#2-wrap) {}; %
  \node[plate caption=#2-wrap] (#2-caption) {#4}; %
  \node[plate=(#2-wrap)(#2-caption), #1] (#2) {}; %
}

\makeatother
\usetikzlibrary{backgrounds}
\usetikzlibrary{calc,quotes,arrows.meta}
\usetikzlibrary{shapes.misc,positioning,calc,graphs,arrows}

\pgfdeclarelayer{edgelayer}
\pgfdeclarelayer{nodelayer}
\pgfsetlayers{edgelayer,nodelayer,main}

\definecolor{hexcolor0xbfbfbf}{rgb}{0.749,0.749,0.749}

\tikzset{>=latex}
\tikzstyle{none}   = [inner sep=0pt]
\tikzstyle{line}   = [ -, thick, shorten <=1pt, shorten >=1pt ]
\tikzstyle{arrow}  = [ ->, thick, shorten <=1pt, shorten >=1pt ]
\tikzstyle{ardash} = [ dashed, ->, thick, shorten <=1pt, shorten >=1pt ]

\tikzstyle{box} = [rectangle, minimum width=1.5cm, minimum height=1.5cm,text centered, draw=black, inner sep=7pt]
\tikzstyle{neuron} = [circle, minimum width=4mm, very thick, draw=blue!80!black]

\tikzstyle{empty}=[circle,opacity=0.0,text opacity=1.0,inner sep=0pt]
\tikzstyle{box}=[rectangle,fill=White,draw=Black]
\tikzstyle{filled}=[circle,thick,fill=hexcolor0xbfbfbf,draw=Black]
\tikzstyle{hollow}=[circle,thick,fill=White,draw=Black]
\tikzstyle{param}=[rectangle,fill=Black,draw=Black,inner sep=0pt,minimum width=4pt,minimum height=4pt]
\tikzstyle{paramhollow}=[rectangle,thick,fill=White,draw=Black,inner sep=0pt,minimum
width=4pt,minimum height=4pt]

\usepackage{pgfplots}                               %
\pgfplotsset{compat=newest}
\pgfplotsset{plot coordinates/math parser=false}
\usepgfplotslibrary{statistics}
\newlength\figureheight
\newlength\figurewidth
\newlength\figureheightsmall
\newlength\figurewidthsmall
\definecolor{POSTcolor}{rgb}{0.48, 0.20, 0.58} %
\definecolor{Qcolor}{rgb}{0.00, 0.53, 0.22} %

\makeatletter
\tikzset{
  prefix after node/.style={
    prefix after command={\pgfextra{#1}}
  },
  /semifill/ang/.store in=\semi@ang,
  /semifill/ang=0,
  semifill/.style={
    circle, draw,
    prefix after node={
      \typeout{aaa \semi@ang}
      \let\nodename\tikz@last@fig@name
      \fill[/semifill/.cd, /semifill/.search also={/tikz}, #1]
        let \p1 = (\nodename.north), \p2 = (\nodename.center) in
        let \n1 = {\y1 - \y2} in
        (\nodename.\semi@ang) arc [radius=\n1, start angle=\semi@ang, delta angle=180];
    },
  }
}
\makeatother

\DeclareRobustCommand{\sd}[1]{\color{black!80!white}\scriptstyle #1}

\usepackage[capitalize,noabbrev]{cleveref}

\theoremstyle{plain}

\theoremstyle{definition}

\theoremstyle{remark}

\newcommand{\x}{\boldsymbol{x}}
\newcommand{\y}{\boldsymbol{y}}
\newcommand{\m}{\boldsymbol{m}}
\newcommand{\tb}{\boldsymbol{t}}
\newcommand{\ub}{\boldsymbol{u}}
\newcommand{\vb}{\boldsymbol{v}}
\newcommand{\wb}{\boldsymbol{w}}
\newcommand{\z}{\boldsymbol{z}}
\newcommand{\s}{\boldsymbol{s}}
\newcommand{\f}{\boldsymbol{f}}
\renewcommand{\L}{\mathcal{L}}
\newcommand{\Y}{\mathcal{Y}}
\newcommand{\N}{\mathcal{N}}
\newcommand{\X}{\mathcal{X}}

\newcommand{\Z}{\mathcal{Z}}
\newcommand{\G}{\mathcal{G}}
\newcommand{\GP}{\mathcal{GP}}

\DeclareMathOperator*{\E}{\mathbb{E}}

\usepackage[textsize=tiny]{todonotes}

\icmltitlerunning{Latent variable model for high-dimensional point process with structured missingness}

\begin{document}

\twocolumn[
\icmltitle{Latent variable model for high-dimensional point process\\ with structured missingness}

\begin{icmlauthorlist}
\icmlauthor{Maksim Sinelnikov}{aalto}
\icmlauthor{Manuel Haussmann}{aalto,sdu}
\icmlauthor{Harri L\"ahdesm\"aki}{aalto}
\end{icmlauthorlist}

\icmlaffiliation{aalto}{Department of Computer Science, Aalto University, Espoo, Finland}
\icmlaffiliation{sdu}{Department of Mathematics and Computer Science, University of Southern Denmark, Odense, Denmark}

\icmlcorrespondingauthor{Maksim Sinelnikov}{maksim.sinelnikov@aalto.fi}

\icmlkeywords{Machine Learning, ICML}

\vskip 0.3in
]

\printAffiliationsAndNotice{}

\begin{abstract}
Longitudinal data are important in numerous fields, such as healthcare, sociology, and seismology, but real-world datasets present notable challenges for practitioners because they can be high-dimensional, contain structured missingness patterns, and measurement time points can be governed by an unknown stochastic process.  While various solutions have been suggested, the majority of them have been designed to account for only one of these challenges. In this work, we propose a flexible and efficient latent-variable model that is capable of addressing all these limitations.  Our approach utilizes Gaussian processes to capture temporal correlations between samples and their associated missingness masks as well as to model the underlying point process. We construct our model as a variational autoencoder together with deep neural network parameterised encoder and decoder models and develop a scalable amortised variational inference approach for efficient model training. We demonstrate competitive performance using both simulated and real datasets. 
\end{abstract}

\section{Introduction}\label{sec:intro}

Longitudinal data arise in many domains such as healthcare, sociology and seismology \citep{longitudinal_applications}. These datasets consist of repeated measurements of unique instances, e.g.\ patients, collected over time. However, real-world applications pose several challenges for practitioners: measurements are typically high-dimensional and contain non-trivial missingness patterns, and time points of the observations are not deterministic but rather arise from an unknown stochastic process. These challenges are characteristic of many real biomedical datasets, such as electronic health records.

Variational autoencoders~(VAEs) have become a popular approach to model high-dimensional data \citep{kingma_aevb, rezende}. However, a notable limitation of standard VAEs is their assumption that the latent variables factorize across samples, hence ignoring correlations between observations and making the models inappropriate for temporal and longitudinal datasets. Several recent works \citep{casale,fortuin,siddhart} have addressed this issue by incorporating Gaussian process (GP) priors for these latent variables, creating a probabilistic model that is capable of modelling arbitrary correlations between latent encodings.

The simplest form of missingness is missing completely at random (MCAR), i.e., the missingness pattern is independent of the observed and unobserved data. While it is generally feasible to handle MCAR in most latent-variable models, more complex patterns of structured missingness~\citep{rubin,structured_missingness} require additional modeling capacities.
Extending VAEs to be able to model various structured missingness patterns has recently become the focus of several papers \citep{collier,ipsen,pmlr-v130-ghalebikesabi21a}. 
However, these methods still lack the ability to model correlations among observations or missingness masks, thus limiting their applicability to temporal data. 

While VAEs have been applied to various biomedical datasets, the existing methods cannot consider observation time points as random variables. Instead, they have to treat time as a deterministic covariate and, therefore, loose useful information embedded in its stochastic nature. A separate line of research has proposed methods to model unknown temporal point processes primarily using GP-based methods \citep{lloyd, liu}. Overall, the field lacks versatile modeling methods that would allow modeling high-dimensional, marked point-processes that may be corrupted by structured missingness.

\begin{figure}[t]
    \centering
\includegraphics[width=0.98\columnwidth]{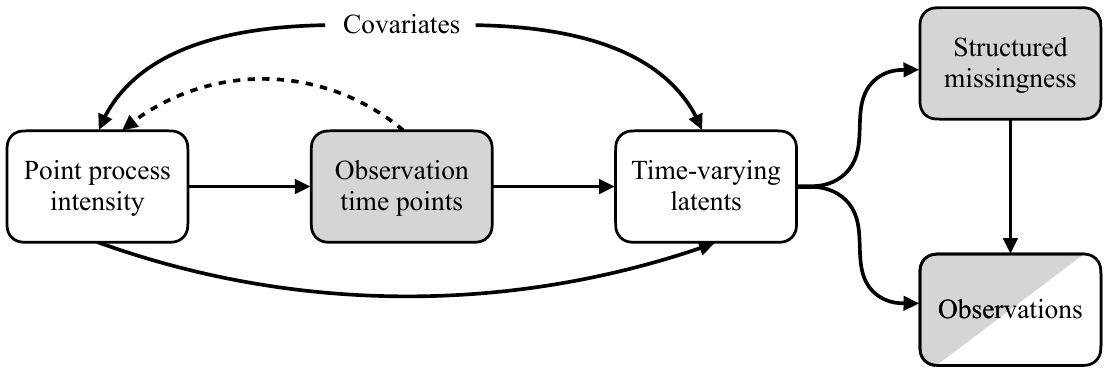}
    \caption{Conceptual overview of our model. Shaded, partially-shaded, and blank rectangles refer to observed, partially observed, and latent components. Dashed arrow corresponds to the dependence of the point process intensity on the previous time points.}  
\label{fig:conceptual_figure}
\end{figure}

\paragraph{Contributions.} 
In this work, we propose a novel deep latent variable model (DLVM), 
that is specifically designed to capture structured missingness and uses temporal point processes to model time. We construct the model by introducing three sets of latent variables with GP priors, which model observations, missingness masks and point process. To adapt the model for longitudinal data, we rely on longitudinal additive kernels \citep{siddhart} for the latent representations of observations and masks. Additionally, to use the information embedded in the temporal point process, we provide the intensity of the point process as an additional input to the GP kernel functions of observations and missingness masks. 
See \cref{fig:conceptual_figure} for a high-level summary of our model and \cref{appsec:supplementary_figures} for an extended visualization. 
We present two variations of our proposal, the simplified \emph{longitudinal latent variable model with structured missingness} (LLSM) without a temporal point process, and the full model, \emph{longitudinal latent variable model for high-dimensional point process with structured missingness} (LLPPSM). 
To summarize, our contributions are that
\begin{itemize}
    \item[\emph{(i)}] we present a latent variable model able to capture structured missingness in the context of longitudinal data;
    \item[\emph{(ii)}] we extend this model by a temporal point process and use the inferred intensity of the process as an additional input to the model;
    \item[\emph{(iii)}] we compare the performance of our two model variants against baseline methods on several datasets and report
state-of-the-art results on a variety of tasks.
\end{itemize}

\section{Related Work}\label{sec:related}

We summarize and compare previous methods in~\cref{tab:summary_table}.

\begin{table*}
\centering
\caption{A summary of related methods.}
\adjustbox{max width=0.98\textwidth}{
\begin{tabular}{lcccccc} 
 \toprule
 Model & Temporal structure & Other covariates & Structured missingness &  Modelling timestamps & Generative & Reference\\ \midrule
 VAE missing & \textcolor{red}{\xmark} & \textcolor{red}{\xmark} & \textcolor{green}{\cmark} & \textcolor{red}{\xmark} & \textcolor{green}{\cmark} & \citet{collier} \\
 not-MIWAE & \textcolor{red}{\xmark} & \textcolor{red}{\xmark} & \textcolor{green}{\cmark} & \textcolor{red}{\xmark} & \textcolor{green}{\cmark} & \citet{ipsen}\\
 GRUI-GAN & \textcolor{green}{\cmark} & \textcolor{red}{\xmark} & \textcolor{green}{\cmark} & \textcolor{red}{\xmark} & \textcolor{green}{\cmark} & \citet{luo}\\
 GPPVAE & \textcolor{green}{\cmark} & Limited & \textcolor{red}{\xmark} & \textcolor{red}{\xmark} & \textcolor{green}{\cmark} & \citet{casale}\\
 GP-VAE & \textcolor{green}{\cmark} & \textcolor{red}{\xmark} & \textcolor{red}{\xmark} & \textcolor{red}{\xmark} & \textcolor{green}{\cmark} & \citet{fortuin}\\
  L-VAE & \textcolor{green}{\cmark} & \textcolor{green}{\cmark} & \textcolor{red}{\xmark} & \textcolor{red}{\xmark} & \textcolor{green}{\cmark} & \citet{siddhart}\\ 
  GPRPP & \textcolor{green}{\cmark} & \textcolor{red}{\xmark} & \textcolor{red}{\xmark} & \textcolor{green}{\cmark} & Limited to timestamps & \citet{liu}\\\midrule
  LLSM & \textcolor{green}{\cmark} & \textcolor{green}{\cmark} & \textcolor{green}{\cmark} & \textcolor{red}{\xmark} & \textcolor{green}{\cmark} & This work\\
  LLPPSM & \textcolor{green}{\cmark} & \textcolor{green}{\cmark} & \textcolor{green}{\cmark} & \textcolor{green}{\cmark} & \textcolor{green}{\cmark} & This work\\
 \bottomrule
 \end{tabular}}
 \label{tab:summary_table}
\end{table*}

\paragraph{Challenges with missing data.}
In his pioneering work, \citet{rubin} identified three classes of missing data: missing completely at random (MCAR), missing at random (MAR), and missing not at random (MNAR). For MCAR, the missingness mechanism is independent of both observed and unobserved variables. In case of MAR, the missingness depends on the observed attributes. Whereas, if data is MNAR, missing readings are dependent on the unobserved data, or systematic factors that are not accounted for in the experiment. The last two are examples of structured missingness. Although MCAR case can be handled by simply excluding missing elements from the analysis without introducing a bias, the same does not hold for the other scenarios. Despite the utility of Rubin's taxonomies, \citet{structured_missingness} emphasized that they do not fully account for high-dimensional patterns of structured missingness, frequently encountered in modern ML applications. They also proposed a set of current grand challenges in learning from data with structured missingness and claimed that the field of learning with missing values needs to be further advanced.

\paragraph{DLVMs for missing data.}
Various methods have been proposed to address the challenge of missing values within generative models. 
\citet{collier} employed a variational autoencoder by concatenating the input with a missingness mask. While this approach can model MAR and MNAR scenarios, it fails to model temporal correlations and does not take into account auxiliary covariate information. 
\citet{mattei} used importance sampling to derive a Missing Importance Weighted Autoencoder (MIWAE) bound for training DLVMs under MAR condition.  Building upon this work, \citet{ipsen} expanded this method to MNAR scenario by directly modeling the missingness mask. However, both these approaches lack the ability to model temporal correlation in the latent space, hence they are not suitable for longitudinal setting.

\paragraph{Temporal DLVMs.} 
To model temporal data, a set of methodologies has emerged, exploring the use of GP priors for latent variables. 
\citet{casale} introduced the GPPVAE model to integrate both view and object information in a GP prior through the product kernel.
\citet{fortuin} proposed the GP-VAE that assigns individual GP prior for the time-series of each subject. 
While these methods allow to model subject-specific temporal structure, they have limited or no functionality to account for possible other auxiliary covariates.
\citet{siddhart} introduced L-VAE, a model that  uses a multi-output additive GP prior and is well-suited for longitudinal data by leveraging carefully designed interaction kernels. The main drawback of all these approaches is their limitation to MCAR modeling which is na\"ive in many domains, such as healthcare.

Another direction of research to deal with temporal data focuses on recurrent neural networks (RNNs). \citet{che} developed GRU-D which incorporates \emph{masking} and a \emph{time interval} into a deep model architecture, making it possible to model structured missingness patterns. They also proposed to use a decaying mechanism to handle irregularly-sampled timestamps. \citet{luo} extended this model for time-series imputation by employing generative adversarial networks (GRUI-GAN). However, it is not straightforward to incorporate auxiliary information into these models.

\paragraph{Longitudinal data analysis.}
An additional line of research that gained popularity in recent years refers to modeling of longitudinal data. Several works have focused on addressing the dependence between timestamps and longitudinal observations in order to avoid bias during the inference \citep{longitudinal_overview, xu2022biascorrection, sang2022functional}. However, although previous methods take into account auxiliary covariate information, they have two limitations. First, they are not applicable for the high-dimensional setting considered in our work as these previous methods were derived for one-dimensional case and employ purely statistical techniques. Second, to the best of our knowledge, the previous methods do not assume missing data mechanisms to depend on timestamps.

\paragraph{Temporal Point Process.} 
Modelling temporal point processes has been a subject of several studies in recent years. Classical statistical approaches \citep{classical_tpp} use maximum likelihood estimation to infer the parameters of a model. 
For this, they require specifying a parametric form of the intensity function which significantly limits their applications. Neural network-based models typically employ RNNs \citep{recurrent_tpp}. 
\citet{lloyd} proposed to model intensity function with Gaussian processes and a squared link function that leads to closed-form solution. \citet{ti} extended this model to variational Fourier features. Both approaches use the current time point, and don't take the previous history of events into account which limits them to model only inhomogeneous Poisson processes, a potentially unrealistic assumption in real-world scenarios. \citet{liu} overcome this problem by incorporating the previous $D$ timestamps into the computation of the GP kernel function.

\section{Methods}\label{sec:method}

\subsection{Background}\label{sec:background}

\paragraph{Problem setup.}
We are given $N = \sum_{p = 1}^{P} n_p$ observations, where $P$ denotes the number of unique instances, e.g., patients, and $n_p$ is the number of observations of instance $p$. 
The longitudinal \emph{response variables} (or \emph{marks}) of instance~$p$ are denoted as $\y_p = [y_1^p ,\ldots, y_{n_p}^p] \in \mathbb{R}^{K \times n_p}$, where each sample $y_i^p \in \Y = \mathbb{R}^K$. 
Each sample $y_i \in \Y$ can be split into observed and missing parts, $y_{i}^{\text{o}}$ and $y_{i}^{\text{m}}$, with a corresponding \emph{binary mask} $m_i \in \{0,1\}^K$ specifying which features of $y_i$ are missing (1 is observed, 0 is missing). 
The \emph{auxiliary covariate} information of instance $p$ is $\x_p = [x_1^p ,\ldots, x_{n_p}^p]$, where each $x_i^p \in \X = \X_1 \times \cdots \times \X_Q$ is a $Q$-dimensional vector. Covariates can be both discrete and continuous and represent, for instance, a patient's age, their gender, etc. 
For notational convenience, we separately denote the measurement \emph{time points} of the subject $p$ as $\tb^p = [t_1^p,\ldots,t_{n_p}^p]^T$ modeled as random variables, and  $\X^{\text{static}}$ as the set of covariates that do not depend on time, such that $t \subset \X$ and $\X^{\text{static}} \subset \X$. 
Observations from all $P$ instances, e.g., patients, form our longitudinal data matrix $\y$, matrix of missing values $\m$, covariate matrix $\x$, and vector of timestamps $\tb$, defined respectively as 
\begin{alignat*}{2}
    \y &= [\y_1,\ldots, \y_P] = [y_1,\ldots, y_N] && \in \mathbb{R}^{K \times N},\\
    \m &= [\m_1,\ldots, \m_P] = [m_1,\ldots, m_N] && \in \{0,1\}^{K \times N},\\
    \x &= [\x_1,\ldots, \x_P] = [x_1,\ldots, x_N] && (\text{size } Q \times N),\\
    \tb &= [\tb_1^T,\ldots, \tb_P^T]^T = [t_1,\ldots, t_N]^T && \in \mathbb{R}^{N}.
\end{alignat*}
We rely on a latent space ${\Z = \mathbb{R}^L}$ and combine the latent embedding for all $N$ samples as ${\z = [z_1,\ldots, z_N] \in \mathbb{R}^{L \times N}}$.

\paragraph{Variational autoencoders.}
Assuming a deep latent variable model $p_{\omega}(y,z) = p_{\psi}(y|z)p_{\theta}(z)$, inference of the posterior $p_{\omega}(z|y) = p_{\psi}(y|z)p_{\theta}(z)/p_{\omega}(y)$ is in general intractable, as the evidence $p_{\omega}(y)$ cannot be computed analytically due to the highly non-linear relationship between $z$ and $y$.
Common practice is to rely on amortized inference \citep{kingma_aevb,rezende}, where a parameterized approximation, $q_{\phi}(\z|\y)$, to the true posterior is inferred by optimizing a lower bound to the evidence,
\begin{equation*}
\log p_{\omega}(\y) \geq {\E}_q\left[\log p_{\psi}(\y|\z)\right] - \text{KL}(q_{\phi}(\z|\y)||p_{\theta}(\z)),
\end{equation*}
with respect to all parameters. Usually, likelihood $p_{\psi}(\y|\z)$, prior $p_{\theta}(\z)$, and variational posterior $q_{\phi}(\z|\y)$ are assumed to be mean-field, i.e., to factorize over their respective random variables.

\paragraph{GP-prior variational autoencoder.}
Despite the computational efficiency provided by a factorized prior $p_{\theta}(\z)$ over the latent variables, its major limitation is the inability to model correlations between data samples. Prior work  addressed this by combining VAEs with GPs, creating a powerful probabilistic model for this task \citep{casale,fortuin,siddhart}.
The key difference is that the factorized prior $p_{\theta}(\z)$ is replaced by a  GP-prior $p_{\theta}(\z | \x)$ which depends on auxiliary information~$\x$. The conditional generative model is then given as
\begin{equation*}
    p_{\omega}(\y | \x) = \int{\prod_{i = 1}^{N}{p_{\psi}(y_i | z_i)} p_{\theta}(\z | \x) d\z}.
\end{equation*}
Defining a mapping from the covariates to the latent space, $\textbf{f} : \X \rightarrow \Z$, such that  $z = \textbf{f}(x) = [f_1(x),\ldots, f_{L}(x)]^T$, these models assume a GP-prior over each $f_l$, 
\begin{equation*}
    f_l(x) \sim \GP\left(\big.\mu_l(x), k_l(x, x' | \theta_l)\right), 
\end{equation*}
where $\mu_l(x)$ is the mean function and $k_l(x, x' | \theta_l)$ is the covariance function parameterized by $\theta_l$.

Given this prior, the $l$-th latent variable ${\bar{\z}_l = f_l(\x) = [f_l(x_1),\ldots, f_l(x_N)]^T}$ follows a multivariate Gaussian distribution across the $N$ data samples
\begin{equation*}
    p_{\theta_l}(\bar{\z}_l | \x) = p_{\theta_l}(f_l(\x)) =  \N\Big(\bar{\z}_l \big| \boldsymbol{0}, K^{(l)}_{\x \x}\Big), 
\end{equation*}
where $K^{(l)}_{\x \x}$ is a $N \times N$ covariance matrix such that $\{K^{(l)}_{\x \x}\}_{ij} = k_l(x_i, x_j | \theta_l)$. We follow common practice and factorize the GP-prior across its $L$ dimensions, such that the conditional prior is given as 
\begin{equation*}
    p_{\theta}(\z | \x) = \prod_{l = 1}^{L}{p_{\theta_l}(\bar{\z}_l | \x)} = \prod_{l = 1}^{L}{\N\left(\bar{\z}_l \big| \boldsymbol{0}, K^{(l)}_{\x \x}\right)}.
\end{equation*}

The main distinction among previous GP-prior models lies in the choice of covariance functions. GPPVAE \citep{casale} and GP-VAE \citep{fortuin} both rely on restricted kernels that do not adequately model longitudinal data structure. Instead, we adopt the proposal by \citet{siddhart} who introduce a flexible additive kernel structure that is specifically designed for longitudinal data and is capable of employing various interactions between continuous and categorical covariates
\begin{equation*}
    k_l(x, x' | \theta_l) = \sum_{r = 1}^{R}{k_{l,r}(x, x' | \theta_{l,r})} + \sigma_{zl}^2,
\end{equation*}
such that 
\begin{equation*}
    p_{\theta}(\z | \x) 
    = \prod_{l = 1}^{L}{\N\left(\bar{\z}_l \Big| \boldsymbol{0}, \sum_{r = 1}^{R} {K^{(l,r)}_{\x \x} + \sigma_{zl}^2 I_{N}}\right)}.
\end{equation*}
\subsection{Modeling structured missingness}
\label{sec:vae_mnar}
These GP-prior models are capable of  dealing with missing values solely by substituting zeros or alternative values and propagating  $y$ through encoder-decoder structure to perform imputation. However, this approach lacks the ability to model specific missingness patterns therefore making them suitable only for an MCAR scenario, an unrealistic assumption in many real applications. 
In this work, we solve this constraint and propose the \emph{longitudinal latent variable model with structured missingness (LLSM)}.
To model non-random missingness in VAE models various approaches exist. For example,  
\citet{mattei} and \citet{ipsen} model the dependency between $y$ and $m$ directly, whereas \citet{collier} propose a VAE model that incorporates an additional latent variable to account for structured missingness. 
In this work, we follow \citet{collier} and introduce a second latent variable~$z^m \in \mathbb {R}^{L_m}$ associated with a missingness mask $m$, while referring to the latent variables associated to $y$ from now on as $z^y \in \mathbb{R}^{L_y}$.  
To properly model MNAR we assume that $m$ depends on both $z^m$ and $z^y$. 
The joint likelihood for a single sample is given as
\begin{align}\label{eq:LLSMjoint}
    p_{\omega}(y^{\text{o}}, z^y, m, z^m | x) 
    &= p_{\psi_{y}}(y^{\text{o}} | z^y, z^m, m)  p_{\psi_{m}}(m | z^y, z^m)\nonumber \\ 
    & \quad \cdot p_{\theta_{y}}(z^y | x)  p_{\theta_{m}}(z^m | x),
\end{align}
where $y^\text{o}$ refers to the observed features specified by $m$.
Also, by optionally conditioning $y$ on $z^m$, we can use any information contained in the missing mask, e.g., during an imputation task. 
To model correlation within the missingness patterns, e.g., across time, or within a patient, we assign $z^m$ a GP prior as well, 
\begin{equation*}
    z^m(x) \sim \GP\left(\big.0, k(x, x' | \theta_m)\right).
\end{equation*}

\begin{figure}
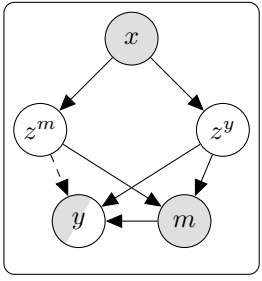

    \centering
    \tikz{
    \node[obs] (x) {$x$};
    \node[latent, below left=of x] (zm) {$z^m$};
    \node[latent, below right=of x] (zy) {$z^y$};
    \node[obs, below right=of zm, xshift=2em] (m) {$m$};
    \node[semifill={gray!25, ang=60}, draw=black,inner sep=1pt, minimum size=20pt, font=\fontsize{10}{10}\selectfont, node distance=1, below left=of zy, xshift=-2em] (y) {$y$};
    \edge{x}{zm,zy};
    \edge[dashed]{zm}{y};
    \edge{zm}{m};
    \edge{zy}{m,y};
    \edge{m}{y};
    \plate{}{(x)(zm)(zy)(m)(y)}{};
    }
    \caption{\emph{The graphical model of LLSM.} Shaded, partially-shaded, and blank circles refer to observed, partially observed, and latent variables. The dashed arrow highlights an optional dependency.}
    \label{fig:DAG_LLSM}
\end{figure}

Additionally, we assume that $y^o$ and $m$ are distributed as
\begin{align*}
    p_{\psi_{y}}(y^{\text{o}} | z^y, z^m, m) &= \N(y | g_{\psi_{y}}(z^y, z^m), \boldsymbol{\sigma}^2) \odot m\\
    p_{\psi_{m}}(m | z^y, z^m) &= \mathcal{B}\textit{er}(m | g_{\psi_{m}}(z^y, z^m)),
\end{align*}
where the decoder functions $g_{\psi_{y}}$ and $g_{\psi_{m}}$ are parameterized by neural networks, $\odot$ denotes an element-wise Hadamard product, and the observational variance parameters ${\boldsymbol{\sigma}^2 = \mathrm{diag}(\sigma^2_{y_1}, \ldots, \sigma^2_{y_K})}$ are optimized jointly with all other parameters via gradient descent. The graphical model of LLSM is shown in \cref{fig:DAG_LLSM}.

We approximate the intractable posterior across all $N$ samples $p(\z^y, \z^m | \x, \y^{\text{o}}, \m)$ using a mean-field amortized inference distribution
\begin{align}
    q_{\phi}(\z^y, \z^m | \y^{\text{o}}, \m) &= q_{\phi_y}(\z^y | \y^{\text{o}})q_{\phi_m}(\z^m | \m) \nonumber \\
    &= \prod_{i = 1}^{N}{q_{\phi_y}(z_{i}^y | y^{\text{o}}_i)q_{\phi_m}(z_{i}^m | m_i)}\label{eq:LLSMq} 
\end{align}
where 
$q_{\phi_y}(z_{i}^y | y^{\text{o}}_i)$ and $q_{\phi_m}(z_{i}^m | m_i)$ are diagonal Gaussian distributions parameterized by neural network-based mappings from the corresponding inputs $y^{\text{o}}_i$ and $m_i$.
The lower bound on the evidence to be optimized is given as 
\begin{align*}
    \log p(\y^o,\m|\x) &\geq 
    {\E}_{q_{\phi}}\left[\log{p_{\psi_y}(\y^{\text{o}}|\z^y, \z^m, \m)}\right] \\ & \quad + {\E}_{q_{\phi}}\left[\log{p_{\psi_m}(\m|\z^y, \z^m)}\right]\\
    &\quad - \hat{\text{KL}}(q_{\phi_y}(\z^y | \y^{\text{o}}) || p_{\theta_y}(\z^y| \x)) \\ 
    &\quad -\hat{\text{KL}}(q_{\phi_m}(\z^m | \m) || p_{\theta_m}(\z^m | \x)),
\end{align*}
where, in order to maintain computational tractability, and to be able to perform mini-batching, we substitute the KL terms with the corresponding upper bounds $\hat{\text{KL}}$ derived by \citet{siddhart} that, for longitudinal data, are tighter than the well-known bound proposed by \citet{titsias}. 
Further details on the lower bound and KL upper bound are given in \cref{appendix:elbo_llsm} and \cref{appendix:kl_scalable}.

\subsection{Modeling Time}

\label{sec:vae_tpp}

Prior work relying on GP-based priors suffers from a second restriction. They often rely on  time $t$ as a primary, or even the only \citep{fortuin}, covariate that is used in the covariance function. 
This implies that the similarity between two measurements $y$ and $y'$ is directly contingent on the corresponding $t$ and $t'$, e.g., their temporal difference when employing a stationary kernel. While this assumption is reasonable in some use cases, it is too limiting to be universally applicable. For instance, consider a scenario where a patient develops a disease, and since the progression varies for each individual, it may be more appropriate for similarity to be determined not solely by the elapsed time since the onset of the disease, but rather by their current well-being, which can be captured by factors such as the frequency of doctor visits. This example highlights a possible bias that may occur if the dependence between timepoints and longitudinal observations is ignored.
To properly account for such variations, we model $t$ by a temporal point process and add the intensity of this point process as an additional input to the GP kernel computation.

\paragraph{Temporal point processes (TPP).}
A TPP \citep{point_process} is a stochastic process over variable-length sequences in some time interval $\mathcal{T} = \left[0, T\right]$ defined via an intensity function $\lambda(t)$. The probability density function of $N$ observed points ${\mathcal{\tb} = \{t_i \in \mathcal{T}\}}$ is defined as
\begin{equation*}
    p(\mathcal{\tb} | \lambda) = \exp\Big(-\int_{\mathcal{T}}{\lambda(t) dt}\Big) \prod_{i = 1}^{N}{\lambda(t_i)}.
\end{equation*}
TPPs can be divided into roughly two classes: inhomogeneous Poisson processes, where the intensity function only depends on the current time point $t$, and self-exciting processes, where the occurrence of events changes the intensity. One type of these self-exciting processes, known as the \emph{Hawkes process} \cite{hawkes}, has an intensity function 
\begin{equation*}
\lambda(t) = \mu + \sum_{t_j < t}{\nu(t - t_j)},
\end{equation*}
where $\nu$ is a \emph{triggering kernel} that characterizes the influence of past events on intensity at time $t$ and $\mu$ is a corresponding baseline. Inspired by the broad applicability of Hawkes processes \citep{hawkes_application}, we adopt the proposal by \citet{liu} to model such self-exciting processes with GPs by computing kernels from the last $D$ timestamps.

\paragraph{GP point processes.} 
Given a latent variable $z^{\lambda}$ with 
\begin{align}\label{eq:latentTPP}
\begin{split}
    z^{\lambda}(t) &\sim \GP\big(0, k_{\theta_{\lambda}}(v_D, {v_D}')\big)\\
    v_D &= t - t_D,
\end{split}
\end{align}
where $t_D$ denotes $D$ previous timestamps before $t$, $v_D$ are the elapsed times between $t$ and $t_D$, and $k_{\theta_{\lambda}}$ is an additive kernel structure, we model the intensity as  
\begin{equation}\label{eq:latentTPP2}
\lambda(t) = (z^{\lambda}(t) + \beta)^2
\end{equation}
where $\beta$ is either a trainable baseline or a function that can depend on static covariates \cite{ti}. We choose a squared link function 
as it provides an analytical tractability \citep{lloyd}.

The posterior distribution of the intensity,
\begin{equation*}
    p(\lambda | \tb) = \frac{p(\lambda)\exp\left(-\int_{\mathcal{T}}{\lambda(t) dt}\right) \prod_{i = 1}^{N}{\lambda(t_i)}}{\int{p(\lambda)\exp\left(-\int_{\mathcal{T}}{\lambda(t)} dt\right) \prod_{i = 1}^{N}{\lambda(t_i)} d\lambda}},
\end{equation*}
is intractable due to the integration over $\lambda$. To overcome this challenge, we approximate it with a variational distribution $p(z^{\lambda} | \boldsymbol{u}) q(\boldsymbol{u})$ that relies on inducing points $\boldsymbol{u}$ for additional scalability \citep{inducing_points}. 
See \cref{appendix:llppsm} for a more detailed discussion.

\begin{figure}
    \centering
    \begin{tikzpicture}
        \node[obs] (t_D) at (0,0) {$t_D$};
        \node[latent, below=of t_D] (z_lambda) {$z^\lambda$};
        \node[latent, right=of z_lambda] (l) {$\lambda$};
        \node[obs, right=of l] (t) {$t$};
        \node[obs, right=of t] (x) {$x$};
        \node[obs, above=of t] (x_static) {$x_p^\text{s}$};
        \node[latent, below left=of x, xshift=-4em] (zm) {$z^m$};
        \node[latent, below right=of l, xshift=4em] (zy) {$z^y$};
        \node[obs, below=of zy] (m) {$m$};
        \node[semifill={gray!25, ang=60}, draw=black,inner sep=2pt, minimum size=20pt, font=\fontsize{10}{10}\selectfont, node distance=1, left=of m] (y) {$y$};
        \edge{t_D}{z_lambda};
        \edge{z_lambda}{l};
        \edge{x_static}{l};
        \edge{l}{t};
        \edge{t}{x};
        \edge{x_static}{x};
        \edge{zy}{m,y};
        \edge{zm}{m};
        \edge[dashed]{zm}{y};
        \edge{m}{y};
        \edge{l}{zm};
        \edge{l}{zy};
        \edge{x}{zm};
        \edge{x}{zy};
        \plate{}{(x_static)(zm)(zy)(y)(t_D)(m)(x)}{};
    \end{tikzpicture}
    \caption{\emph{The graphical model of LLPPSM}. (Partially) shaded, and blank circles refer to (partially) observed, and latent variables. The dashed arrow highlights an optional dependency, $x_p^\text{s}$   are static covariates and $t_D$ the $D$ previous time steps to $t$.} \label{fig:LVAE-MNAR-TPP}
\end{figure}
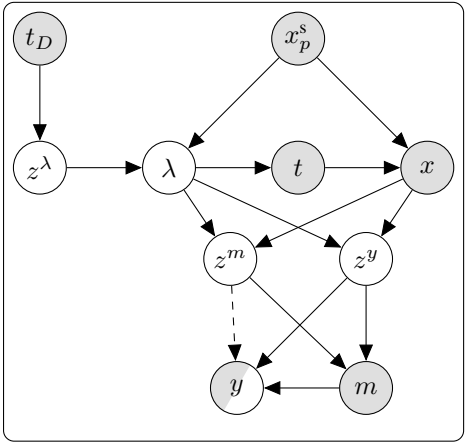

\paragraph{LLPPSM.} 
Combining such a point process with our LLSM model allows us to 
capture intricate missingness patterns and to effectively leverage information embedded in the time points. We call this model \emph{longitudinal latent variable model for high-dimensional point process with structured missingness (LLPPSM)}. See \cref{fig:LVAE-MNAR-TPP} for its plate diagram.

Defining $z^\lambda$ and $\lambda$ as in \cref{eq:latentTPP,eq:latentTPP2}, we extend the GP priors for $z^y$ and $z^m$ by letting their covariance kernel depend on the rate of the TPP $\lambda$ as well, i.e., 
\begin{equation*}
    z^y(x,\lambda) \sim \GP\big(0, k((x,\lambda(t)), (x', \lambda(t'))| \theta_y)\big),
\end{equation*}
and analogously for $z^m$. 
As inference of the full model 
remains intractable, we once again rely on variational inference and use the following variational approximation
\begin{align*}
&q(\z^y,\z^m,z^{\lambda},\boldsymbol{u}|\y^{\text{o}},\m) \\
& \qquad= q_{\phi_y}(\z^y | \y^{\text{o}}) q_{\phi_m}(\z^m | \m) p(z^{\lambda} | \boldsymbol{u}) q(\boldsymbol{u}),
\end{align*}
where $q_{\phi_y}(\z^y | \y^{\text{o}})$ and $q_{\phi_m}(\z^m | \m)$ are defined as in \cref{eq:LLSMq}, and $\boldsymbol{u}$ are the inducing points of $z^{\lambda}$. The bound to be optimized is given as 
\begin{align*}
&\log p(\y^o, \m, \tb |\x^\text{s}) \geq \\
    &\quad {\E}_{q} \left[\log \left(p_{\psi_y}(\y^{\text{o}} \big| \z^y, \z^m, \m)  p_{\psi_m}(\m \big| \z^y,  \z^m) p(\tb \big| \lambda)\right)\right] \\
    &\qquad- \text{KL}(q(\boldsymbol{u}) || p(\boldsymbol{u}))\\ 
    &\qquad- {\E}_{p(z^{\lambda}|\boldsymbol{u}) \cdot q(\boldsymbol{u})} [\hat{\text{KL}}(q_{\phi_y}(\z^y\big| \y^{\text{o}}) || p_{\theta_y}(\z^y \big|\x, \lambda(\tb))]\\ 
    &\qquad- {\E}_{p(z^{\lambda} |\boldsymbol{u}) \cdot q(\boldsymbol{u})}[\hat{\text{KL}}(q_{\phi_m}(\z^m\big| \m) || p_{\theta_m}(\z^m\big|\x, \lambda(\tb))],
\end{align*}
where $\x^\text{s}$ refers to static covariates and $\x$ is composed of $\x^\text{s}$ and $\tb$.
See \cref{appendix:llppsm} for a detailed derivation and discussion.

\subsection{Imputation and future prediction}
Our method can be employed for various tasks such as imputation and future prediction. The imputation is done by substituting missing elements with some intermediate values (in our case zeros), and propagating them through the encoder-decoder structure of our model so that the decoder imputes the missing values. 

Future predictions are obtained by evaluating the posterior predictive distribution $p(y_* \,|\, x_*, \y^{\text{o}}, \x, \m)$ for new data $y_*$ given covariates $x_*$ and all training data. A detailed explanation together with the necessary derivations to approximate this intractable density is given in \cref{appendix:predictive_distribution}.

\subsection{Computational complexity and scalability}
The complexity of LLSM is dominated by computation of KL divergence upper bounds $\hat{\text{KL}}(q_{\phi_y}(\boldsymbol{z}^y | \boldsymbol{y}^{\text{o}}) || p_{\theta_y}(\boldsymbol{z}^y| \boldsymbol{x}))$ and $\hat{\text{KL}}(q_{\phi_m}(\boldsymbol{z}^m | \boldsymbol{m}) || p_{\theta_m}(\boldsymbol{z}^m | \boldsymbol{x}))$, which, by employing the techniques from \citet{siddhart}, have complexity $\mathcal{O}(\sum_{p=1}^{P}{{n_p}^3} + NM^2)$, where $M$ is the number of inducing points. We also adopt the mini-batching scheme from \citet{siddhart} that provides additional scalability to large-sized datasets in terms of memory consumption.

For LLPPSM, an additional complexity comes from the point process computation, which corresponds to $\mathcal{O}(NM^2D^2)$ \citep{liu}, therefore the total complexity is $\mathcal{O}(\sum_{p=1}^{P}{{n_p}^3} + NM^2D^2)$, where $NM^2$ vanishes as $NM^2D^2$ dominates it. When training LLPPSM, we also employ mini-batching in a similar fashion as for LLSM to achieve additional scalability.

\section{Experiments}\label{sec:exp}
We demonstrate the efficiency of our proposal on various tasks, such as missing value imputation, long-term prediction, for synthetic as well as real-world healthcare datasets. 
We compare against a variety of models: GPPVAE \cite{casale} serves as a general GP-prior representative,  L-VAE \cite{siddhart} as a variant specifically designed for longitudinal type of data, GRUI-GAN \cite{luo} is a GAN-based model capable of modelling non-random missingness, and mean imputation/prediction is a common simple baseline. As GRUI-GAN is not designed for generative purposes, we only provide imputation results for this method. 
For each method we evaluate its mean-squared error (MSE) and report the mean performance as well as standard deviation computed over five runs. The lowest mean in each experiment is marked bold in the corresponding table. 
 See \cref{appsec:expsetup} for further experimental details (e.g., hyperparameters, kernel structures) that are not specified in the main text and \cref{appsec:neural_nets} for neural network architectures. An implementation of our proposed methods is available at \url{https://github.com/sinelnikovmaxim/MPP-VAE}.

 \begin{figure*}[!t]
    \centering
\includegraphics[width=0.9\textwidth]{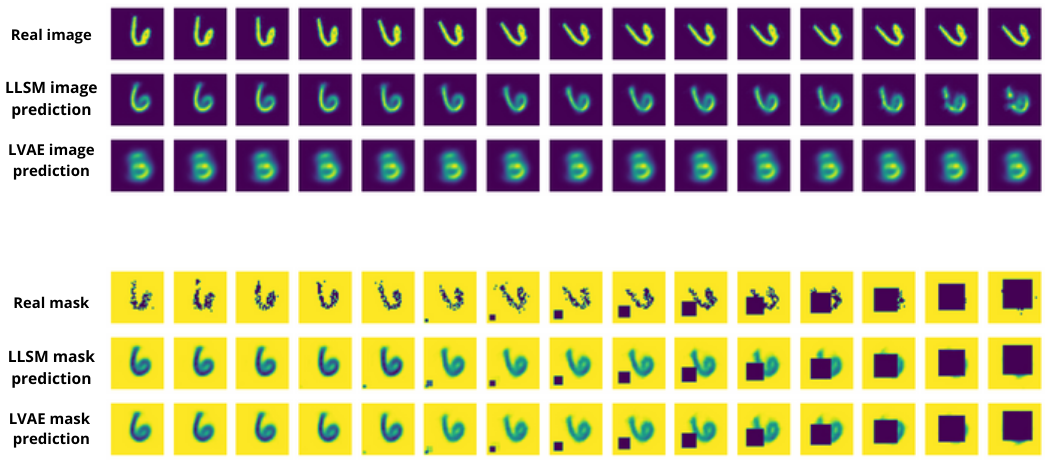}
    \caption{Future predictions of data (top) and missingness mask (bottom) on the regularly sampled health MNIST  dataset. Although predictions of mask look almost identical, the prediction of data cannot be captured by L-VAE, whereas LLSM does it very accurately.}
\label{fig:regular_prediction}
\end{figure*}

\subsection{Regularly sampled Health MNIST}

\begin{table}
\centering
\caption{Imputation MSEs on the regularly sampled health MNIST  dataset. The percentage represents maximum probability of pixel being missing.}
\adjustbox{max width=0.98\columnwidth}{
\begin{tabular}{lccccc} 
 \toprule
 Method & 50\% & 75\% & 90\% \\ \midrule
 mean imputation & $0.266 \sd{\pm 0.000}$ & $0.314 \sd{\pm 0.000}$ & $0.373 \sd{\pm 0.000}$ \\
 GPPVAE & $0.248 \sd{\pm 0.004}$ &  $0.291 \sd{\pm 0.004}$ & $0.379 \sd{\pm 0.015}$ \\
 GRUI-GAN & $0.224  \sd{\pm  0.037}$ & $0.218  \sd{\pm  0.012}$  & $0.269  \sd{\pm  0.028}$ \\
 L-VAE & $\mathbf{0.124 \sd{\pm 0.009}}$ & $0.283 \sd{\pm 0.062}$ & $0.373 \sd{\pm 0.001}$\\
 LLSM (ours) & $\mathbf{0.124  \sd{\pm 0.008}}$ & $\mathbf{0.144 \sd{\pm 0.009}}$ & $\mathbf{0.174 \sd{\pm 0.016}}$\\
 \bottomrule
 \end{tabular}}
\label{tab:regular_imputation}
\end{table}

\begin{table}[!t]
\centering
\caption{Future data prediction MSEs on the regularly sampled health MNIST dataset. The percentage represents maximum probability of pixel being missing.}
\adjustbox{max width=0.98\columnwidth}{
\begin{tabular}{lccccc} 
 \toprule
 Method & 50\% & 75\% & 90\% \\ \midrule
 mean prediction & $0.040 \sd{\pm 0.000}$ &  $0.042 \sd{\pm 0.000}$ & $0.048 \sd{\pm 0.000}$\\
 GPPVAE & $0.041 \sd{\pm 0.000}$ &  $0.041 \sd{\pm 0.000}$ & $0.048 \sd{\pm 0.001}$ \\
 L-VAE & $\mathbf{0.021 \sd{\pm 0.002}}$ &  $0.038 \sd{\pm 0.008}$ & $0.047 \sd{\pm 0.000}$ \\
 LLSM (ours) & $0.023 \sd{\pm 0.002}$ & $\mathbf{0.024 \sd{\pm 0.001}}$ & $\mathbf{0.026 \sd{\pm 0.002}}$ \\
 \bottomrule
 \end{tabular}}
 \label{tab:regular_prediction_data}
\end{table}

To simulate a high-dimensional longitudinal dataset with structured missingness, we used a modified version of the MNIST dataset \citep{mnist} called Health MNIST \citep{healthmnist}. We chose two digits, `3’ and `6’, to represent two biological genders. We simulated a shared time-related effect by shifting all digit instances towards the right corner over time. In our experiments, half of the instances remain
healthy and half get a disease. To demonstrate changes in the laboratory measurements of the diseased individuals, we rotated digits with the amount of rotation depending on the time from disease diagnosis. Each sample has in total five covariates: \emph{time}, \emph{id}, \emph{diseasePresence}, \emph{diseaseTime}, and \emph{gender}, where \emph{id} serves as the identifier of a specific instance. The timestamps of all observations are regularly sampled which is why we only evaluated LLSM. 

To model MNAR, the probability of each pixel being missing depends on
the color intensity of that pixel: the higher the intensity, the higher is the probability of the pixel being unobserved. For MAR, we
applied a square-shaped missingness mask to the images. If a patient is healthy, no box is applied. For diseased patients, a mask is only applied upon the onset of the disease. 
Afterwards, the mask starts to increase in size
linearly as time progresses. See \cref{fig:regular_prediction} for an illustration. 
In this case, missingness depends on both unobserved signal as well as on covariate information.

The training set consists of $P = 900$ unique instances, each having $n_p = 20$ time points. The test set contains $100$ unique instances, with $15$ last observations for each instance. When performing future prediction, the model conditions on all training data as well as first five observations of each instance, that are kept separately.

\cref{tab:regular_imputation} shows that our method outperforms all baselines on the task of missingness imputation.
The same holds for future prediction of data from covariates (\cref{tab:regular_prediction_data}), with the exception for the simplest missingness scenario where L-VAE is slightly better. We also performed future prediction of missingness mask by the same approach. Because none of the baseline methods is able to model the mask $m$ explicitly, we separately modelled missingness by training an L-VAE with a Bernoulli likelihood for $m$. 
The results are shown in \cref{tab:regular_prediction_mask} and show that the mask prediction is almost identical except for the case with the highest missingness when LLSM is slightly better. 
In \cref{fig:regular_prediction}, we demonstrate visually the benefits of our model for future prediction for the case of 75$\%$ maximum probability of missingness. 

\begin{table}[!t]
\centering
\caption{Future missingness prediction MSEs on the regularly sampled health MNIST dataset. The percentage represents maximum probability of pixel being missing.}
\adjustbox{max width=0.98\columnwidth}{
\begin{tabular}{lccccc} 
 \toprule
 Method & 50\%  & 75\% & 90\% \\ \midrule
 L-VAE & $\mathbf{0.032 \sd{\pm 0.000}}$ & $\mathbf{0.038 \sd{\pm 0.002}}$ & $0.040 \sd{\pm 0.002}$  \\
 LLSM (ours) & $\mathbf{0.031 \sd{\pm 0.002}}$ & $\mathbf{0.038 \sd{\pm 0.003}}$ & $\mathbf{0.038 \sd{\pm 0.002}}$ \\
 \bottomrule
 \end{tabular}}
 \label{tab:regular_prediction_mask}
\end{table}

\subsection{Irregularly sampled Health MNIST}
\begin{table}[!t]
\centering
\caption{Imputation MSEs on the irregularly sampled health MNIST dataset. The percentage represents maximum probability of pixel being missing.}
\adjustbox{max width=0.98\columnwidth}{
\begin{tabular}{lccccc} 
 \toprule
 Method & 50\% & 75\% & 90\% \\ \midrule
 mean imputation & $0.259 \sd{\pm 0.000}$ & $0.335 \sd{\pm 0.000}$ & $0.380 \sd{\pm 0.000}$ \\ 
 GPPVAE & $0.239 \sd{\pm 0.002}$ &  $0.319 \sd{\pm 0.001}$ & $0.396 \sd{\pm 0.004}$ \\
 GRUI-GAN & $0.165 \sd{\pm 0.016}$ & $0.203 \sd{\pm 0.021}$ & $0.277 \sd{\pm 0.035}$\\
 L-VAE & $0.171 \sd{\pm 0.045}$ & $0.264 \sd{\pm 0.059}$ & $0.379 \sd{\pm 0.001}$ \\
 LLSM (ours) & $0.130  \sd{\pm 0.007}$ & $0.163 \sd{\pm 0.011}$ & $\mathbf{0.191 \sd{\pm 0.012}}$ \\
 LLPPSM (ours) & $\mathbf{0.128 \sd{\pm 0.005}}$ &  $\mathbf{0.162 \sd{\pm 0.007}}$ & $0.207 \sd{\pm 0.028}$ \\
 \bottomrule
 \end{tabular}}
 \label{tab:irregular_imputation}
\end{table}

\begin{figure*}[!t]
    \centering
\includegraphics[width=0.9\textwidth]{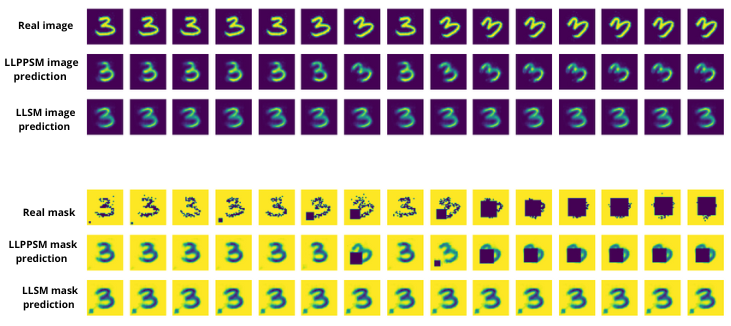}
    \caption{Future predictions of data (top) and missingness mask (bottom) on irregularly sampled health MNIST  dataset.}  
\label{fig:irregular_prediction}
\end{figure*}

We modified the previous setup such that timepoints come from a random process and similarity in the observations depends on covariates as well as the underlying rate of the TPP that was discussed in \cref{sec:vae_tpp}.
We implemented it in a following way: for healthy patients, the timestamps come from a homogeneous Poisson process with intensity $\lambda = 0.1$ and the digit is not modified, whereas for diseased patients, timestamps are generated according to a Hawkes process, with baseline intensity $\mu = 0.5$ and the digit rotation depends on the intensity of process at the moment: the higher the intensity, the stronger the rotation. For healthy patients we modelled MNAR as in the previous setup, while for diseased patients we also applied a square-shaped mask with its size being proportional to the intensity of the process at that moment.
\cref{tab:irregular_imputation} shows that both of our models improve upon the baselines in all imputation scenarios. Moreover, Tables~\ref{tab:irregular_prediction_data} and~\ref{tab:irregular_prediction_mask} show that LLPPSM outperforms LLSM in both future data and mask prediction tasks. The \cref{fig:irregular_prediction} depicts that although LLSM is capable of capturing the general form of an image, it cannot model the rotation properly due to the limited kernel component related to time whereas LLPPSM does it well. The same holds for prediction of mask. The inferred mean intensity function of the point process for one individual can be found in \cref{appsec:supplementary_figures}.

\subsection{Physionet data}
We evaluated our model on healthcare data from the
2012 Physionet Challenge \citep{physionet}. The dataset contains around 12,000 patients monitored on the
intensive care unit (ICU) for 48 hours. We modelled measurements of 37 different attributes, such as glucose level, heart rate, body temperature, etc. The dataset is extremely sparse, with about 85$\%$ missing values.  We cannot directly measure imputation performance due to the lack of ground truth data for missing values, hence, to test the learned representations of our models, we used this dataset only for future prediction. We predicted values of laboratory attributes given the knowledge of the first ten measurements for a patient in the test set. As auxiliary covariates, we employed the following variables: \emph{time of the measurement}, \emph{id}, \emph{type of ICU}, \emph{gender}, and \emph{in-hospital mortality}. More information regarding Physionet data can be found in \cref{appsubsec:physionet}. We present the results in \cref{tab:physionet}. LLSM performs best, with LLPPSM performing worse. This can be explained by the fact that many observations are taken regularly each hour, making the temporal process be pseudo-stochastic, which is reflected in the intensity of the point process that starts to explode at each hour timepoint, hence modelling it just brings additional redundant information to the model. 

\begin{table}[!t]
\centering
\caption{Future data prediction MSEs on the irregularly sampled health MNIST dataset. The percentage represents maximum probability of pixel being missing.}
\adjustbox{max width=0.98\columnwidth}{
\begin{tabular}{lccccc} 
 \toprule
 Method & 50\% & 75\% & 90\% \\ \midrule
 mean prediction & $0.039 \sd{\pm 0.000}$ & $0.044 \sd{\pm 0.000}$ & $0.047 \sd{\pm 0.000}$ \\
 GPPVAE & $0.040 \sd{\pm 0.000}$ &  $0.044 \sd{\pm 0.001}$ & $0.049 \sd{\pm 0.000}$ \\
 L-VAE & $0.029 \sd{\pm 0.005}$ &  $0.036 \sd{\pm 0.006}$ & $0.047 \sd{\pm 0.000}$ \\
 LLSM (ours) & $0.030 \sd{\pm 0.001}$ & $0.031 \sd{\pm 0.001}$ & $0.031 \sd{\pm 0.001}$ \\
 LLPPSM (ours) & $\mathbf{0.025 \sd{\pm 0.002}}$ & $\mathbf{0.027 \sd{\pm 0.001}}$ & $\mathbf{0.030 \sd{\pm0.003}}$ \\
 \bottomrule
 \end{tabular}}
 \label{tab:irregular_prediction_data}
\end{table}

\begin{table}[!t]
\centering
\caption{Future missingness prediction MSEs on the irregularly sampled health MNIST dataset. The percentage represents maximum probability of pixel being missing.}
\adjustbox{max width=0.98\columnwidth}{
\begin{tabular}{lccccc} 
 \toprule
 Method & 50\%  & 75\%  & 90\% \\ \midrule
 L-VAE & $0.063 \sd{\pm 0.000}$ & $0.067 \sd{\pm 0.001}$ & $0.068 \sd{\pm 0.001}$  \\
 LLSM (ours) & $0.063 \sd{\pm 0.002}$ & $0.067 \sd{\pm 0.001}$ & $0.066 \sd{\pm 0.001}$ \\
 LLPPSM (ours) & $\mathbf{0.054 \sd{\pm 0.006}}$ & $\mathbf{0.056 \sd{\pm 0.005}}$ & $\mathbf{0.058 \sd{\pm 0.005}}$ \\
 \bottomrule
 \end{tabular}}
 \label{tab:irregular_prediction_mask}
\end{table}

\begin{table}[!t]
\centering
\caption{Future prediction on Physionet dataset.}
\adjustbox{max width=0.98\textwidth}{
\begin{tabular}{lc} 
 \toprule
 Method   & MSE\\ \midrule
 mean  prediction  & $0.785 \sd{\pm 0.000}$ \\
 GPPVAE  & $0.786 \sd{\pm 0.001}$ \\
 L-VAE  & $0.720 \sd{\pm 0.008}$ \\
 LLSM (ours)  & $\mathbf{0.713  \sd{\pm 0.006}}$ \\
 LLPPSM (ours)  & $0.745  \sd{\pm 0.009}$ \\
 \bottomrule
 \end{tabular}}
\label{tab:physionet}
\end{table}

\section{Conclusions}
In this work, we introduced a novel probabilistic framework for multivariate data with missing values. First, we developed a deep latent variable model, LLSM, that models structured missingness via separate set of latent variables. Second, we extended this model by utilizing temporal point process to account for stochastic nature of timepoints, LLPPSM. Our methods are specifically designed for longitudinal type of data by leveraging GPs to define priors for latent variables. We demonstrated excellent performance of both models on different representation learning tasks and expect them to become useful tools in the analysis of high-dimensional temporal and longitudinal data.

\section*{Acknowledgements}
We acknowledge the computational resources provided by the Aalto Science-IT project. This work was supported by the Research Council of Finland (decision number: 359135). 

\section*{Impact statement}
This paper presents a work whose goal is to introduce a new method for analyzing high-dimensional data with missing values in the longitudinal scenario. We do not see any potential harmful societal consequences of our work.

\bibliography{example_paper}
\bibliographystyle{icml2024}

\newpage
\appendix
\onecolumn

\begin{center}
    \Huge \textsc{Appendix}
\end{center}

\section{Deriving the ELBO for LLSM}

\label{appendix:elbo_llsm}

By introducing a variational distribution $q_{\phi}(\z^y, \z^m | \y^{\text{o}}, \m)$, the marginal likelihood can be expanded as
\begin{align*}
    \log p_{\omega}(\y^{\text{o}},\m | \x) &= \iint{q_{\phi}(\z^y, \z^m | \y^{\text{o}}, \m) \log {\frac{p_{\omega}(\z^y, \z^m | \y^{\text{o}},\m) p_{\omega}(\y^{\text{o}},\m |\x)} {p_{\omega}(\z^y, \z^m | \y^{\text{o}},\m)}} d\z^y d\z^m} \\
    &= \iint{q_{\phi}(\z^y, \z^m | \y^{\text{o}}, \m) \log {\frac{p_{\omega}(\y^\text{o}, \m, \z^y, \z^m | \x)} {p_{\omega}(\z^y, \z^m | \y^{\text{o}},\m)}} d\z^y d\z^m} \\
    &= \iint{q_{\phi}(\z^y, \z^m | \y^{\text{o}}, \m) \log {\frac{p_{\psi}(\y^{\text{o}}, \m |\z^y, \z^m) p_{\theta}(\z^y, \z^m |\x)} {p_{\omega}(\z^y, \z^m |\y^{\text{o}},\m)}} d\z^y d\z^m} \\
    &= \iint{q_{\phi}(\z^y, \z^m | \y^{\text{o}}, \m) \log {p_{\psi}(\y^{\text{o}}, \m |\z^y, \z^m)} d\z^y d\z^m} \\
    &\qquad+ \iint{q_{\phi}(\z^y, \z^m | \y^{\text{o}}, \m) \log \frac{p_{\theta}(\z^y, \z^m | \x)} {p_{\omega}(\z^y, \z^m | \y^{\text{o}},\m)} d\z^y d\z^m} \\
    &= {\E}_{q_{\phi}}\left[\log {p_{\psi}(\y^{\text{o}}, \m |\z^y, \z^m)}\right]
     \\
     &\qquad+ \iint{q_{\phi}(\z^y, \z^m | \y^{\text{o}}, \m) \log \frac{p_{\theta}(\z^y, \z^m | \x) q_{\phi}(\z^y, \z^m | \y^{\text{o}}, \m)} {p_{\omega}(\z^y, \z^m | \y^{\text{o}},\m) q_{\phi}(\z^y, \z^m | \y^{\text{o}}, \m)} d\z^y d\z^m} \\
    &= {\E}_{q_{\phi}}\left[\log {p_{\psi}(\y^{\text{o}}, \m |\z^y, \z^m)}\right]\\
    &\qquad+ \iint{q_{\phi}(\z^y, \z^m | \y^{\text{o}}, \m) \log \frac{q_{\phi}(\z^y, \z^m | \y^{\text{o}}, \m)} {p_{\omega}(\z^y, \z^m | \y^{\text{o}},\m)} d\z^y d\z^m}  \\
    &\qquad- \iint{q_{\phi}(\z^y, \z^m | \y^{\text{o}}, \m) \log \frac{q_{\phi}(\z^y, \z^m | \y^{\text{o}}, \m)}{p_{\theta}(\z^y, \z^m | \x)} d\z^y d\z^m} \\
    &= {\E}_{q_{\phi}}\left[\log {p_{\psi}(\y^{\text{o}}, \m |\z^y, \z^m)}\right] \\
    &\qquad+ \text{KL}(q_{\phi}(
    \z^y, \z^m | \y^{\text{o}}, \m) || p_{\omega}(\z^y, \z^m | \y^{\text{o}},\m)) \\
    &\qquad - \text{KL}(q_{\phi}(\z^y, \z^m | \y^{\text{o}}, \m) || p_{\theta}(\z^y, \z^m | \x)).
\end{align*}
Hence, 
\begin{align*}
    &\log p_{\omega}(\y^{\text{o}},\m | \x) -  \text{KL}(q_{\phi}(\z^y, \z^m | \y^{\text{o}}, \m) || p_{\omega}(\z^y, \z^m | \y^{\text{o}},\m)) =  \\
    &\qquad {\E}_{q_{\phi}}\left[\log {p_{\psi}(\y^{\text{o}}, \m |\z^y, \z^m)}\right] 
    - \text{KL}(q_{\phi}(\z^y, \z^m | \y^{\text{o}}, \m) || p_{\theta}(\z^y, \z^m |\x)).
\end{align*}
As the KL divergence term is positive, we get
\begin{align*}
    \log p_{\omega}(\y^{\text{o}},\m | \x) &\geq {\E}_{q_{\phi}}\left[\log {p_{\psi}(\y^{\text{o}}, \m |\z^y, \z^m)}\right] - \text{KL}(q_{\phi}(\z^y, \z^m | \y^{\text{o}}, \m) || p_{\theta}(\z^y, \z^m | \x))\\
    &= \mathcal{L}(\phi, \psi, \theta; \y^{\text{o}}, \m).
\end{align*}

By assuming the following factorizations:
\begin{align*}
    p_{\psi}(\y^{\text{o}}, \m |\z^y, \z^m) &= p_{\psi_y}(\y^{\text{o}}|\z^y, \z^m, \m) p_{\psi_m}(\m |\z^y, \z^m) \\
    q_{\phi}(\z^y, \z^m | \y^{\text{o}}, \m) &= q_{\phi_y}(\z^y| \y^{\text{o}}) q_{\phi_m}(\z^m| \m)\\
    p_{\theta}(\z^y, \z^m | \x) &= p_{\theta_y}(\z^y| \x) p_{\theta_m}(\z^m| \x),
\end{align*}
the ELBO simplifies to
 \begin{align*}
    \mathcal{L}(\phi, \psi, \theta; \y^{\text{o}}, \m) &= {\E}_{q_{\phi}}\left[\log {p_{\psi_y}(\y^{\text{o}}|\z^y, \z^m, \m)}\right] + {\E}_{q_{\phi}} \left[\log {p_{\psi_m}(\m|\z^y, \z^m)}\right]\\
    &\qquad-  \text{KL}(q_{\phi_y}(\z^y | \y^{\text{o}}) || p_{\theta_y}(\z^y| \x)) -  \text{KL}(q_{\phi_m}(\z^m | \m) || p_{\theta_m}(\z^m | \x)),
\end{align*}

where $\psi_y$, $\psi_m$, $\phi_y$, $\phi_m$ are parameterized by neural networks and $q_y$, $q_m$ are corresponding variational distributions of $\z^y$ and $\z^m$.

By approximating the KL terms with the corresponding upper bounds $\hat{{\text{KL}}}$ that are necessary for computational scalability (see \cref{appendix:kl_scalable}), the final ELBO is given as
 \begin{align*}
    \mathcal{L}(\phi, \psi, \theta; \y^{\text{o}}, \m) &\geq {\E}_{q_{\phi}}\left[\log {p_{\psi_y}(\y^{\text{o}}|\z^y, \z^m, \m)}\right] + {\E}_{q_{\phi}} \left[\log {p_{\psi_m}(\m|\z^y, \z^m)}\right]\\
    &\qquad- \hat{{\text{KL}}}(q_{\phi_y}(\z^y | \y^{\text{o}}) || p_{\theta_y}(\z^y| \x)) - \hat{\text{KL}}(q_{\phi_m}(\z^m | \m) || p_{\theta_m}(\z^m | \x)).
\end{align*}

\section{Scalable KL Divergence Computation}

\label{appendix:kl_scalable}

Here we review the KL upper bound from \citet{siddhart} that implements a scalable KL divergence computation. 
Optimising the variational objective requires us to evaluate $L$ KL terms $\text{KL}({\N(\bar{\boldsymbol{\mu_l}}, W_l) || \N(\boldsymbol{0}, \Sigma_l))}$,
where $\bar{\boldsymbol{\mu_l}} = [\mu_{\phi, l}(y_1) ,\ldots, \mu_{\phi, l}(y_N)]^T,  W_l = \mathrm{diag}(\sigma^2_{\phi, l}(y_1), \ldots, \sigma^2_{\phi, l}(y_N))$, and $\Sigma_l = \sum_{r = 1}^{R} {K^{(r, l)}_{\x\x} + \sigma_{zl}^2 I_{N}}$. For notational convenience, we drop  the index $l$. The exact computation has $\mathcal{O}(N^3)$ complexity, making it impractical for large datasets. Therefore, instead of computing it, we will use an upper bound to the KL that comes from the fact that any lower bound for the prior GP marginal log-likelihood induces an upper bound to the KL divergence. \citet{titsias} proposed the free-form variational lower bound for a GP marginal log-likelihood $\log \N(\z \big |\boldsymbol{0}, \Sigma)$ by assuming a set of $M$ inducing locations $\s = [s_1,\ldots,s_M]$ in $\X$, with the corresponding inducing function values $\boldsymbol{u} = [f(s_1),\ldots, f(s_M)]^T = [u_1,\ldots, u_M]^T$, such that
\begin{align*}
    p(\boldsymbol{u}) &= \N(\boldsymbol{u} \big| \boldsymbol{0}, K_{\s\s})\\
    p(\f \big| \boldsymbol{u}) &= \N(\f \big| K_{\x\s}K^{-1}_{\s\s} \boldsymbol{u}, \tilde{K}), \\
    \tilde{K} &= K_{\x\x} -K_{\x\s}K^{-1}_{\s\s}K_{\s\x}\\
    p(\z \big|\f) &= \N(\z \big|\f, \sigma^2_z I_N),
\end{align*}

and the corresponding lower bound is $\L(\z; \Sigma) = \N(\z \big| \boldsymbol{0}, K_{\x\s}K^{-1}_{\s\s}K_{\s\x} + \sigma^2_z I_N) - \frac{1}{2 \sigma^2_z} \text{tr}(\tilde{K})$, where $\text{tr}(\cdot)$ is the matrix trace. Titsias bound is known to be tight when $M$
 is large enough and the covariance function is smooth enough. Longitudinal data, however, always contain categorical covariates corresponding to instance \emph{ids}, making the covariance function necessarily non-continuous. 

 To still get a tighter bound, we separate the additive component that corresponds to the instance \emph{id} from the other covariates, so that covariance function has the following form $\Sigma = K^{(A)}_{\x\x} + \hat{\Sigma}$, where $\hat{\Sigma} = \mathrm{diag}(\hat{\Sigma}_1,\ldots, \hat{\Sigma}_P)$, $\hat{\Sigma}_p = K^{(R)}_{\x_p\x_p} + \sigma^2_z I_{n_p}$ contains all terms with instance-specific \emph{id} and $K^{(A)}_{\x\x} = \sum_{r = 1}^{R-1} {K^{(r)}_{\x\x}}$ contains the other components. \citet{siddhart} proposed the following upper bound for KL
 \begin{align*}
    \text{KL} \leq \frac{1}{2}\left(\text{tr}\left(\bar{\Sigma}^{-1}W\right) + \bar{\boldsymbol{\mu}}^T\bar{\Sigma}^{-1}\bar{\boldsymbol{\mu}} - N + \log |\bar{\Sigma}| - \log|W| + \sum_{p = 1}^{P}{\text{tr}\left(\hat{\Sigma}^{-1}_{p} \tilde{K}^{(A)}_{\x_p\x_p}\right)}\right),
 \end{align*}

where $\bar{\Sigma} = K^{(A)}_{\x\s}K^{{(A)}^{-1}}_{\s\s}K^{(A)}_{\s\x} + \hat{\Sigma}$ and $\tilde{K}^{(A)}_{\x_p\x_p} = K^{(A)}_{\x_p\x_p} - K^{(A)}_{\x_p\s}K^{{(A)}^{-1}}_{\s\s}K^{(A)}_{\s\x_p}$. This bound has a computational complexity $\mathcal{O}(\sum_{p=1}^{P}{{n_p}^3 + NM^2)}$ and \citet{siddhart} proved it to be tighter than the corresponding bound by Titsias for longitudinal datasets.
Despite the reduced complexity, a problem still remains. For every gradient descent step, the algorithm has to iterate through the entire dataset, requiring a substantial allocation of memory and computational time. This issue can be solved using a similar technique as the one presented by \citet{hensman}, with adaptation to the properties of a longitudinal kernel. 
We will only present the final bound and refer the reader to \citet{siddhart} for a detailed derivation.
Defining $I_{p_i}$ to be the index of the $i$th sample for the $p$th patient and $\bar{\boldsymbol{\mu}}_p = \left[\bar{\boldsymbol{\mu}}_{I_{p_i}},\ldots, \bar{\boldsymbol{\mu}}_{I_{n_p}}\right]^T$ to be a a sub-vector of $\bar{\boldsymbol{\mu}}$ that is related to the $p$th patient, the unbiased estimate of the KL divergence upper bound, computed from the batch with instances $\mathrm{P}_\text{batch} \subset{\{1,\ldots,P\}}$, is defined as
\begin{align*}
\hat{\text{KL}} &= \frac{1}{2} \frac{P}{|\mathrm{P}_\text{batch}|}\sum_{p \in \mathrm{P}} \left(\left(K^{(A)}_{\x_p\s}K^{{(A)}^{-1}}_{\s\s} \boldsymbol{m}_{H} - \bar{\boldsymbol{\mu}}_p\right)^T {\hat{\Sigma}^{-1}}_p \left(K^{(A)}_{\x_p\s}K^{{(A)}^{-1}}_{\s\s} \boldsymbol{m}_{H} - \bar{\boldsymbol{\mu}}_p\right) + \sum_{i = 1}^{n_p} ({\hat{\Sigma}}^{-1}_p)_{ii} \sigma^2_{\phi}(\y_{I_{p_i}}) \right. \\
&\qquad +\left. \quad \log |\hat{\Sigma}_p| + \text{tr}\left(\hat{\Sigma}^{-1}_p {\tilde{K}^{(A)}}_{\x_p\x_p}\right) + \text{tr}\left(\left(K^{{(A)}^{-1}}_{\s\s} H K^{{(A)}^{-1}}_{\s\s}\right)\left(K^{(A)}_{\s\x_p}\hat{\Sigma}^{-1}_p K^{(A)}_{\x_p\s}\right)\right) \right.
\left. - \sum_{i = 1}^{n_p} \log \sigma^2_{\phi}(\y_{I_{p_i}}) \right)\\
&\qquad-\frac{N}{2} + \text{KL}\left[\mathcal{N}(\boldsymbol{m}_{H}, H) || \mathcal{N}(\boldsymbol{0}, K^{(A)}_{\s\s})\right],
\end{align*}
where $\m_H$ and $H$ are variational parameters computed via natural gradients. 

\section{Various LLPPSM specifications}
\label{appendix:llppsm}

We define three sets of latent variables: $z^y, z^m$ and $z^\lambda$. Observations $\y$, masks $\m$ and timestamps $\tb$ are modelled as random variables. The complete joint probability is given as:
\begin{align*}
    p_{\omega}(\y^{\text{o}}, \z^y, \m, \z^m, \tb, z^{\lambda} | \x^\text{s}, \vb_D) 
    = &p_{\psi_{y}}(\y^{\text{o}} | \z^y, \z^m, \m)  p_{\psi_{m}}(\m | \z^y, \z^m) p_{\theta_{y}}(\z^y | \x, \lambda(\tb))  p_{\theta_{m}}(\z^m | \x, \lambda(\tb))\\
    &\qquad p(\tb | \lambda) p_{\theta_{\lambda}}(z^{\lambda} | \vb_D),
\end{align*}
where $\vb_D$ corresponds to elapsed times between $\tb$ and $D$ previous timepoints that occurred before $\tb$, denoted as $\tb_D$, $\x^\text{s}$ refers to static covariates and $\x$ consists of $\tb$ and $\x^\text{s}$. To compute the marginal likelihood, we have to marginalize over the latent variables as
\begin{align*}   
 p_{\omega}(\y^{\text{o}}, \m, \tb | \x^\text{s}, \vb_D) &= \iiint{ p_{\omega}(\y^{\text{o}}, \z^y, \m, \z^m, \tb, z^{\lambda} | \x^\text{s}, \vb_D) d\z^y d\z^m dz^{\lambda}}\\
&=\iiint{p_{\psi_{y}}(\y^{\text{o}} | \z^y, \z^m, \m)  p_{\psi_{m}}(\m | \z^y, \z^m) p_{\theta_{y}}(\z^y | \x, \lambda(\tb))  p_{\theta_{m}}(\z^m | \x, \lambda(\tb))}\\
&\qquad \cdot p(\tb | \lambda) p_{\theta_{\lambda}}(z^{\lambda} | \vb_D) d\z^y d\z^m dz^{\lambda}.
\end{align*}
Factorizing the joint likelihood w.r.t.\ the observation given the latent variables, we get
    \begin{align*}
    p_{\omega}(\y^{\text{o}}, \m, \tb | \x^\text{s}, \vb_D) = \iiint&\prod_{i = 1}^{N}{p_{\psi_{y}}(y^{\text{o}}_i | z_{i}^y, z_{i}^m, m_i)  p_{\psi_{m}}(m_i | z_{i}^y, z_{i}^m)}
     p_{\theta_{y}}(\z^y | \x, \lambda(\tb)) p_{\theta_{m}}(\z^m | \x, \lambda(\tb))\\
     &\qquad p(\tb | \lambda) p_{\theta_{\lambda}} (z^{\lambda} | \vb_D) d\z^y d\z^m dz^{\lambda}.
    \end{align*}
    
We assume the following Gaussian process priors
\begin{align*}
    z^{\lambda}(t) &\sim{\G}P\left(\big.0, k(v_D, {v_D}' | \theta_{\lambda})\right)\\
    z^y(x,\lambda) &\sim {\G}P\left(\big.0, k((x,\lambda(t)), (x', \lambda(t'))| \theta_y)\right)\\
    z^m(x,\lambda) &\sim {\G}P\left(\big.0, k((x,\lambda(t)), (x', \lambda(t'))| \theta_m)\right).
\end{align*}
The decoders of $\y$ and $\m$ are parameterized by neural networks that predict the mean of the generative distributions
\begin{align*}
    p_{\psi_{y}}(y^{\text{o}}_i | z_{i}^y, z_{i}^m, m_i) &= N(y_i \big| g_{\psi_{y}}(z_{i}^y, z_{i}^m), \text{diag}(\sigma^2_{y_1}, \ldots, \sigma^2_{y_K})) \odot m_i\\
    p_{\psi_{m}}(m_i \big| z_{i}^y, z_{i}^m) &= Ber(m_i|g_{\psi_{m}}(z_{i}^y, z_{i}^m)),
\end{align*}
where $\sigma^2_{y_1}, \ldots, \sigma^2_{y_K}$ are jointly optimised via gradient descent, and $\odot$ is the elementwise Hadamard product.

The likelihood of the temporal point process for instance $p$ is given as 
\begin{align*}
    p(\tb_p | \lambda) &= \exp\Big(-\int_{\mathcal{T}}{\lambda(t) dt}\Big) \prod_{i = 1}^{n_p}{\lambda(t_i)},\\
\text{where}\qquad \lambda(t) &= (z^{\lambda}(t) + \beta)^2,
\end{align*}
and $\beta$ is a learnable offset.
    
The covariances of the GPs of $z^y$ and $z^m$ are parameterized by the additive kernels discussed in \cref{sec:method}. For $z^{\lambda}$, we instead rely on the following additive structure:
\begin{align*}
    k(v_D, {v_D}' | \theta_{\lambda}) &= \sum_{d=1}^{D}{\mathbbm{1}({{v_d}}) \mathbbm{1}({{{v_d}'}}) \cdot \gamma_d \cdot \exp\left(-\frac{(v_d - {v_d}^{'})^2} {2 l_d^2}\right)}, \\
    \mathbbm{1}({{v_d}}) &= \begin{cases} 
      1, \text{if} \ {v_d} < \infty \\
      0,\text{otherwise}\\
   \end{cases},
\end{align*}
where $v_d$ is the elapsed time between $t$ and $d$th timepoint, $t_d$, that happened before $t$,  and infinite, if there is no available information about the past event. Hence, the above kernel depends on past events and on the current time point. 

As the posterior inference is not tractable, we rely on variational inference and choose the following approximation to the posterior 
\begin{equation*}
q(\z^y, \z^m, z^{\lambda}, \boldsymbol{u} \big| \y^{\text{o}}, \m) = q_{\phi_y}(\z^y | \y^{\text{o}}) \cdot q_{\phi_m}(\z^m | \m) \cdot p(z^{\lambda} | \boldsymbol{u}) \cdot q(\boldsymbol{u}),    
\end{equation*}
where we also employed inducing points of $z^{\lambda}$, denoted by $\boldsymbol{u}$ with
$q(\boldsymbol{u}) = N(m_{\lambda}, S)$. Note that these inducing points are different from those of the scalable KL bound in \cref{appendix:kl_scalable}. 
To shorten the notation we will denote this variational posterior as $q$ in the following derivations.
The ELBO is given as
\begin{align*}
     \mathcal{L}& = {\E}_{q}\left[\log \frac{p_{\omega}(\y^{\text{o}}, \m, \tb, \z^y, \z^m, z^{\lambda}, \ub| \x^\text{s}, \vb_D)} {q_{\phi_y}(\z^y | \y^{\text{o}}) \cdot q_{\phi_m}(\z^m | \m) \cdot p(z^{\lambda}| \boldsymbol{u}) \cdot q(\boldsymbol{u})}\right]\\
    &= {\E}_{q} \left[\log \left(p_{\psi_y}(\y^{\text{o}} | \z^y, \z^m, \m)  p_{\psi_m}(\m | \z^y,  \z^m) p(\tb | \lambda)\right)\right] - {\E}_{q} \left[\log (\frac{q(\boldsymbol{u})}{p(\boldsymbol{u})})\right] \\
    &\qquad - {\E}_{q} \left[\log (\frac{q_{\phi_y}(\z^y | \y^{\text{o}})}{p(\z^y | \x, \lambda(\tb))})\right]
     - {\E}_{q}\left[\log (\frac{q_{\phi_m}(\z^m | \m)}{p(\z^m | \x, \lambda(\tb))})\right]\\\\
     &= {\E}_{q} \left[\log \left(p_{\psi_y}(\y^{\text{o}} | \z^y, \z^m, \m)  p_{\psi_m}(\m | \z^y,  \z^m) p(\tb | \lambda)\right)\right] - \text{KL}(q(\boldsymbol{u}) || p(\boldsymbol{u})) \\\\ 
     &\qquad- {\E}_{p(z^{\lambda} | \boldsymbol{u}) \cdot q(\boldsymbol{u})} \left[\text{KL}(q_{\phi_y}(\z^y | \y^{\text{o}}) || p(\z^y | \x, \lambda(\tb))\right] - {\E}_{p(z^{\lambda} | \boldsymbol{u}) \cdot q(\boldsymbol{u})} \left[\text{KL}(q_{\phi_m}(\z^m | \m) || p(\z^m | \x, \lambda(\tb))\right]\\\\
     &\ge {\E}_{q} \left[\log \left(p_{\psi_y}(\y^{\text{o}} | \z^y, \z^m, \m)  p_{\psi_m}(\m | \z^y,  \z^m) p(\tb | \lambda)\right)\right] - \text{KL}(q(\boldsymbol{u}) || p(\boldsymbol{u})) \\\\ 
     &\qquad- {\E}_{p(z^{\lambda} | \boldsymbol{u}) \cdot q(\boldsymbol{u})} \left[\hat{\text{KL}}(q_{\phi_y}(\z^y | \y^{\text{o}}) || p(\z^y | \x, \lambda(\tb))\right] - {\E}_{p(z^{\lambda} | \boldsymbol{u}) \cdot q(\boldsymbol{u})} \left[\hat{\text{KL}}(q_{\phi_m}(\z^m | \m) || p(\z^m | \x, \lambda(\tb))\right],
\end{align*}
where $\hat{\text{KL}}$ is the corresponding upper bound for KL divergence from \cref{appendix:kl_scalable} and expectations involving upper bounds are estimated via Monte Carlo sampling.
The generative part can be further decomposed into
\begin{align*}
     &{\E}_{q} \left[\log\left (p_{\psi_y}(\y^{\text{o}} | \z^y, \z^m, \m)  p_{\psi_m}(\m | \z^y,  \z^m) p(\tb | \lambda)\right)\right] \\
     &\qquad= {\E}_{q_{\phi}}{\left[\log \left(p_{\psi_y}(\y^{\text{o}} | \z^y, \z^m, \m)\right)\right]} + {\E}_{q_{\phi}} {\left[\log \left(p_{\psi_m}(\m | \z^y, \z^m)\right)\right]} + {\E}_{p(z^{\lambda} | \ub) \cdot q(\ub)}\left[\log (p(\tb | \lambda)\right],
\end{align*}
where $q_{\phi}(\z^y, \z^m | \y^{\text{o}}, \m) = q_{\phi_y}(\z^y| \y^{\text{o}}) \cdot q_{\phi_m}(\z^m| \m)$.

The expectations involving $\z^y$ and $\z^m$ are computed by sampling from their variational distributions and utilizing the reparameterization trick \citep{kingma_aevb}, whereas expectation involving likelihood of the point process can be evaluated in a closed from due to the chosen squared link function as we show below for the timepoints of individual $p$.

To lighten the notation, we use $\mathcal{L}_T := {\E}_{p(\z^{\lambda} | \ub) \cdot q(\ub)} \left[\log(p(\tb | \lambda)\right]$, drop subscript $p$ denoting the individual and drop subscript $D$ from $v_D$. Also, by $\s$ we denote inducing point locations of $\ub$. First, we integrate out $\ub$:
\begin{align*}
    q(z^{\lambda}) &= \int_{U}{p(z^{\lambda} | \ub) \cdot q(\ub) d\ub} = N(z^{\lambda}\big|\tilde{\mu}, \tilde{\Sigma}), \\
\text{where}\qquad \tilde{\mu}(v) &= k_{v \s} K_{\s \s}^{-1} m_{\lambda},\\
    \tilde{\Sigma}(v, {v}') &= K_{v {v}'} - k_{v \s} K_{\s \s}^{-1} k_{\s {v}'} + k_{v \s} K_{\s \s}^{-1} S K_{\s \s}^{-1} k_{\s {v}'},
\end{align*}

and $U$ is a space of inducing values.
We write $\mathcal{L}_T$ as
\begin{equation*}
    \mathcal{L}_T = \sum_{n}{{\E}_{q(z^{\lambda})}\left[\log\lambda(t_n)\right]} -{\E}_{q(z^{\lambda})}\Big[\int_{T}{\lambda(t) dt}\Big] = \sum_{n}{{\E}_{q(z^{\lambda})}\left[\log\left({z^{\lambda}(t_n) + \beta}\right)^2\right]} - \underbrace{{\E}_{q(z^{\lambda})}\Big[\int_{T}{\lambda(t) dt}\Big]}_{:=\mathcal L_t}, 
\end{equation*}
where we sum over all timepoints of the individual.
Via \citet{lloyd}, we have 
\begin{align*}
{\E}_{q(z^{\lambda})}\left[\log\left({z^{\lambda}(t_n) + \beta}\right)^2\right] &= \int_{-\infty}^{\infty}{\log ({(z^{\lambda}(t_n) + \beta})^2) N(z^{\lambda}(t_n)\big|\tilde{\mu}, \tilde{\sigma}^2) dz^{\lambda}(t_n)} \\
    &= -\tilde{G}\Big(-\frac{({\tilde{\mu} + \beta})^2} {2 \tilde{\sigma}^2}\Big) + \log\Big(\frac{\tilde{\sigma}^2}{2}\Big) - C,
\end{align*}
where $\tilde{\sigma}^2$ is the diagonal element of $\tilde{\Sigma}$,  $C$ is the Euler-Mascheroni constant and $\tilde{G}$ is the confluent hyper-geometric function.

Following the derivations of \citet{liu}, we compute $\mathcal{L}_t$ as:
\begin{align*}
    \mathcal{L}_t &= {\E}_{q(z^{\lambda})}\left[\int_{T}{\lambda(t) dt}\right] = {\E}_{q(z^{\lambda})}\left[\int_{T}{(z^{\lambda}(t) + \beta)^2 dt}\right] \\
    &= \int_{T}{{\E}_{q(z^{\lambda})}\left[(z^{\lambda}(t) + \beta)^2\right] dt} \\
    &= \int_{T}{\left({\E}_{q(z^{\lambda})}\left[(z^{\lambda}(t)^2\right]  + 2\beta {\E}_{q(z^{\lambda})}\left[z^{\lambda}(t)\right] + \beta^2\right)dt} \\
    &= \int_{T}{{\E}_{q(z^{\lambda})}\left[z^{\lambda}(t)\right]^2 dt} + \int_{T}{\text{Var}_{q(z^{\lambda})}\left[z^{\lambda}(t)\right] dt} + 2\beta \int_{T}{{\E}_{q(z^{\lambda})}\left[z^{\lambda}(t)\right] dt} + \beta^2 |T|  \\
    &= \sum_{n}{\left[\int_{t_{n-1}}^{t_n}{{\E}_{q(z^{\lambda})}\left[z^{\lambda}(t)\right]^2 dt} + \int_{t_{n-1}}^{t_n}{\text{Var}_{q(z^{\lambda})}\left[z^{\lambda}(t)\right] dt} + 2\beta \int_{t_{n-1}}^{t_n}{{\E}_{q(z^{\lambda})}\left[z^{\lambda}(t)\right] dt}\right] + \beta^2 |T|}.
\end{align*}
Each integral is computed as follows:
\begin{align*}
    \int_{\tb_{n-1}}^{\tb_n}{{\E}_{q(z^{\lambda})}\left[z^{\lambda}(t)\right]^2 dt} &= m_{\lambda}^T K_{\s \s}^{-1} \Psi_n K_{\s \s}^{-1} m_{\lambda}, \\
    \int_{t_{n-1}}^{t_n}{\text{Var}_{q(z^{\lambda})}\left[z^{\lambda}(t)\right] dt} &= \sum_{d=1}^{D}{\gamma_d \int_{t_{n-1}}^{t_n}{\mathbbm{1}}({v_d})dt} - \text{tr}(K_{\s \s}^{-1} \Psi_n) + \text{tr}(K_{\s \s}^{-1}SK_{\s\s}^{-1} \Psi_n), \\
    \int_{t_{n-1}}^{t_n}{{\E}_{q(z^{\lambda})}\left[z^{\lambda}(t)\right] dt} &= \Phi_n(\s)^T K_{\s \s}^{-1} m_{\lambda}, \\
    \Psi_n(\s, \s') &= \int_{t_{n-1}}^{t_n}{K(\s, v(t))K(v(t), \s') dt},\\
    \Phi_n(\s) &= \int_{t_{n-1}}^{t_n}{K(\s, v(t)) dt}.
\end{align*}
$\Phi$ and $\Psi$ each have closed form solutions which we obtain by evaluating the integrals for the sum of SE kernels.
\begin{align*}
    \Phi_n(\s) &= \sum_{d=1}^{D}{\mathbbm{1}(\s_d) \mathbbm{1}(v_{d_{n}}) \gamma_d \frac{\sqrt{\pi l_d^2}}{\sqrt{2}} \left[\text{erf}\left(\frac{{v_d}_n - \s_d}{\sqrt{2 l_d^2}}\right) - \text{erf}\left(\frac{{v_d}_{n-1} - \s_d}{\sqrt{2 l_d^2}}\right)\right]}, \\
    \Psi_n(\s, \s') &= \sum_{i,j = 1}^{D}\mathbbm{1}(\s_i) \mathbbm{1}(\s_j') \mathbbm{1}({v_i}_{n}) \mathbbm{1}({v_j}_{n}) \gamma_i \gamma_j \frac{\sqrt{\pi l_i^2 l_j^2}}{\sqrt{2 \cdot (l_i^2 + l_j^2)}} \exp\left(-\frac{(\s_i + {t_i}_{n} - \s_j' - {t_j}_{n})^2}{2(l_i^2 + l_j^2)}\right) \\
&\qquad\left[\text{erf}\left(\frac{l_i^2({v_j}_n - \s_j') + l_j^2({v_i}_n - \s_i)}{\sqrt{2 l_i^2 l_j^2(l_i^2 + l_j^2)}}\right) - \text{erf}\left(\frac{l_i^2({v_j}_{n-1} - \s_j') + l_j^2({v_i}_{n-1} - \s_i)}{\sqrt{2 l_i^2 l_j^2(l_i^2 + l_j^2)}}\right)\right],
\end{align*}
where $l_i$, $\gamma_i$ are kernel hyperparameters.

\section{Predictive distribution}
\label{appendix:predictive_distribution}
Given training samples $\y^{\text{o}}$, covariate information $\x$ and masks $\m$, the predictive distribution for a new observation $y_*$, given covariates $x_*$ is
\begin{align*}
p(y_* \,|\, x_*, \y^{\text{o}}, \x, \m) &= \int p(y_* \,|\, z^y_*, z^m_*)p(z^y_*, z^m_* \,|\, x_*, \y^{\text{o}}, \x, \m) \,dz^y_* \,dz^m_* \\
&\approx \int \underbrace{p(y_* \,|\, z^y_*, z^m_*)}_\text{decode GP predictions}\underbrace{p(z^y_* \,|\, x_*, \lambda(t_*), \z^y, \x, \lambda(\tb))}_\text{GP posterior of $z^y_*$}\underbrace{p(z^m_* \,|\, x_*, \lambda(t_*), \z^m, \x, \lambda(\tb))}_\text{GP posterior of $z^m_*$} \\
&\quad \underbrace{q(z^{\lambda}_*)}_\text{variational posterior of $z^{\lambda}_*$} \underbrace{q_{\phi_y}(\z^y \,|\, \y^{\text{o}})}_\text{encode data} \underbrace{q_{\phi_m}(\z^m \,|\, \m)}_\text{encode mask} \,dz^\lambda_* \,dz^y_* \,dz^m_* \,d\z^y \,d\z^m,
\end{align*}
where timestamps $\tb$ and $t_*$ belong to $\x$ and $x_*$, respectively, and $p(z^y_* \,|\, x_*, \lambda(t_*), \z^y, \x, \lambda(\tb))$ and $p(z^m_* \,|\, x_*, \lambda(t_*), \z^m, \x, \lambda(\tb))$ are GP posteriors such that
\begin{align*}
&p(z_* \,|\, x_*, \lambda(t_*), \z, \x, \lambda(\tb)) = \N(\tilde{\mu}, \tilde{\Sigma}),\\
&\tilde{\mu} = {K_{w_* \wb}} (K_{\boldsymbol{w}\boldsymbol{w}} + \sigma_{z}^2 I_{N})^{-1} \z,\\
&\tilde{\Sigma} = K_{w_* w_*} + \sigma_{z}^2 I_{N_*} - {K_{w_* \wb}} (K_{\boldsymbol{w}\boldsymbol{w}}  + \sigma_{z}^2 I_{N})^{-1}{K_{\wb w_*}},
\end{align*}
where $\boldsymbol{w} = \left(\x, \lambda(\tb)\right)$, $w_* = \left(x_*, \lambda(t_*)\right)$ and $N_*$ is a number of elements for prediction. Above, we dropped the superscripts, meaning that the same formulae hold for both $z^y$ and $z^m$ with respect to their kernel hyperparameters. The same approach holds for deriving $p(m_* \,|\, x_*, \y^\text{o}, \x, \m)$. In order to sample from these predictive distributions, ancestral sampling can be employed. Computing the predictive distributions scales cubically for this case. To get the idea of how to implement a scalable predictive distribution using low-rank inducing point approximation, we refer the reader to the derivations by \citet{siddhart}. 

\section{Experimental setups}
\label{appsec:expsetup}
We employed identical kernel structures for the GPs of both $z^y$ and $z^m$ across all of the datasets mentioned below. Nonetheless, it's important to note that, in general, these kernel structures could differ between the two. We also selected sixty inducing points for each GP model for all setups and chose the latent dimension to be 32.

\subsection{HealthMNIST variants}
\label{appsubsec:healthmnist}
For the \emph{Health MNIST regularly sampled} dataset we use the following covariates: \emph{time}, \emph{id}, \emph{diseasePresence}, \emph{diseaseTime} and \emph{gender}. When running LLSM on the corresponding dataset, we relied on the following additive kernel structure
\begin{equation*}
f_{\text{ca}}(\text{id}) + f_{\text{se}}(\text{time}) + f_{\text{ca} \times \text{se}}(\text{id} \times \text{time}) + f_{\text{ca}\times \text{se}} (\text{gender} \times \text{time}) + f_{\text{ca} \times \text{se}} (\text{diseasePresence} \times \text{diseaseTime}),
\end{equation*}
where se denotes squared exponential kernel and ca is referred to categorical one.

For \emph{Health MNIST irregularly sampled} dataset, \emph{time}, \emph{id}, \emph{gender} and \emph{diseasePresence} were used as covariates for LLSM and LLPPSM. In case of LLPPSM, we also added intensity of the point process for covariance computation. When using LLSM, we employed the following kernel structure
\begin{equation*}
f_{\text{ca}}(\text{id}) + f_{\text{se}}(\text{time}) + f_{\text{ca} \times \text{se}}(\text{id} \times \text{time}) + f_{\text{ca}\times \text{se}} (\text{gender} \times \text{time}),
\end{equation*}
whereas for LLPPSM
\begin{equation*}
f_{\text{ca}}(\text{id}) + f_{\text{se}}(\text{time}) + f_{\text{ca} \times \text{se}}(\text{id} \times \text{time}) + f_{\text{ca}\times \text{se}} (\text{gender} \times \text{time}) + f_{\text{ca} \times \text{se}}(\text{id} \times \text{intensity}) + f_{\text{ca}\times \text{se}} (\text{gender} \times \text{intensity}).
\end{equation*}

For both variants, we used the Adam optimiser \citep{adam} as implemented in Pytorch \citep{pytorch}, with a learning rate equal to 0.001, which was selected based on cross-validation. 
After having pretrained a standard VAE, we trained both LLSM and LLPPSM on 1000 epochs, employing early stopping.

When training LLPPSM, we defined separate $\beta$ parameters for ``healthy" and ``sick" instances and optimised them jointly together with other parameters. For the temporal point process part of the model, the number of previous timestamps, $D$, is 15.

\subsection{Physionet dataset}
\label{appsubsec:physionet}
We selected 2000 patients for training, 1917 for validation and performed future prediction on 100 patients, not included in training and validation sets. We used the following covariates: \emph{time}, \emph{id}, \emph{type of ICU}, \emph{gender} and \emph{in-hospital mortality}, with the corresponding additive kernel structure for LLSM
\begin{align*}
f_{\text{ca}}(\text{id}) + f_{\text{se}}(\text{time}) + f_{\text{ca} \times \text{se}}(\text{id} \times \text{time}) + f_{\text{ca}\times \text{se}} (\text{gender} \times \text{time}) \\ 
+ f_{\text{ca}\times \text{se}} (\text{type of ICU} \times \text{time}) + f_{\text{ca}\times \text{se}} (\text{in-hospital mortality} \times \text{time}),
\end{align*}
and for LLPPSM, including intensity as an additional variable
\begin{align*}
&f_{\text{ca}}(\text{id}) + f_{\text{se}}(\text{time}) + f_{\text{ca} \times \text{se}}(\text{id} \times \text{time}) + f_{\text{ca}\times \text{se}} (\text{gender} \times \text{time})
+ f_{\text{ca}\times \text{se}} (\text{type of ICU} \times \text{time})\\ + &f_{\text{ca}\times \text{se}} (\text{in-hospital mortality} \times \text{time}) + f_{\text{ca} \times \text{se}}(\text{id} \times \text{intensity}) + f_{\text{ca}\times \text{se}} (\text{gender} \times \text{intensity})\\
+ &f_{\text{ca}\times \text{se}} (\text{type of ICU} \times \text{intensity}) + f_{\text{ca}\times \text{se}} (\text{in-hospital mortality} \times \text{intensity}) ,
\end{align*}
The optimisation was done by Adam optimiser \citep{adam} using Pytorch \citep{pytorch}, with the learning rate 0.001. Both models were trained for 400 epochs, employing early stopping, after having pretrained them by standard VAE.

When training LLPPSM, we defined separate $\beta$ parameters based on in-hospital mortality attribute and optimised them jointly with other parameters. For temporal point process part, the number of previous timestamps, $D$, is 15.

Moreover, we found it useful to drop the dependence from $z^m$ to $y$ as was discussed in \cref{sec:vae_mnar}. Our intuition is that in this case, if the dependence is left, we are obliged to apply predictive distribution (\cref{appendix:predictive_distribution}) for both $z^y$ and $z^m$, which, due to the highly complex and sparse nature of this dataset, cannot be modelled highly accurately, leading to the accumulation of additional error.

\section{Neural network architectures}
\label{appsec:neural_nets}
\cref{healthmnist_architecture} describes neural network architecture used for both Health MNIST setups that consists of convolutional and feedforward layers. \cref{physionet_architecture} describes neural network architecture for the Physionet dataset. In this case, we employed a  multi layered perceptron (MLP).

\begin{table}
\caption{Neural Network architecture used in Health MNIST variants}
\label{healthmnist_architecture}
\begin{center}
\begin{tabular}{lcc}
\toprule
&  Hyperparameter &  Value   \\
\midrule
 \multirow{12}{*}{Inference network of both $z^y$ and $z^m$}  & Dimensionality of input  & $36 \times 36$    \\
& Number of filters per convolution layer & $144$  \\
& Kernel size & $3 \times 3$  \\
& Stride & $2$  \\
& Pooling & Max pooling \\
& Pooling kernel size & $2 \times 2$\\
& Pooling stride & $2$\\
& Number of feedforward layers & $2$\\
& Width of feedforward layers  & $300, 30$\\
& Dimensionality of latent space & $L$\\
& Activation function of layers & RELU\\
\midrule 
\multirow{8}{*}{Generative network of both $y$ and $m$} & Dimensionality of input  & $L$  \\
& Number of transposed convolution layers  & $3$  \\
& Number of filters per transposed convolution layer  & $256$\\
& Kernel size  & $4 \times 4$  \\
& Stride  & $2$  \\
& Number of feedforward layers  & $2$  \\
& Width of feedforward layers  & $30, 300$  \\
&Activation function of layers & RELU\\
\bottomrule
\end{tabular}
\end{center}
\end{table}

\begin{table}
\caption{Neural Network architecture used in Physionet dataset}
\label{physionet_architecture}
\begin{center}
\begin{tabular}{lcc}
\toprule
&  Hyperparameter &  Value   \\
\midrule
 \multirow{5}{*}{Inference network of both $z^y$ and $z^m$}  & Dimensionality of input  & $37$ \\
& Number of feedforward layers & $2$\\
& Width of feedforward layers & $300, 30$ \\
& Dimensionality of latent space & $L$ \\
& Activation function of layers &  RELU \\
\midrule 
\multirow{4}{*}{Generative network of both $y$ and $m$} & Dimensionality of input & $L$\\
& Number of feedforward layers  & $3$  \\
& Width of feedforward layers  & $30, 30, 300$  \\
&Activation function of layers & RELU\\
\bottomrule
\end{tabular}
\end{center}
\end{table}

\section{Supplementary figures}
\label{appsec:supplementary_figures}

\begin{figure*}[!h]
    \centering
\includegraphics[width=0.8\textwidth]{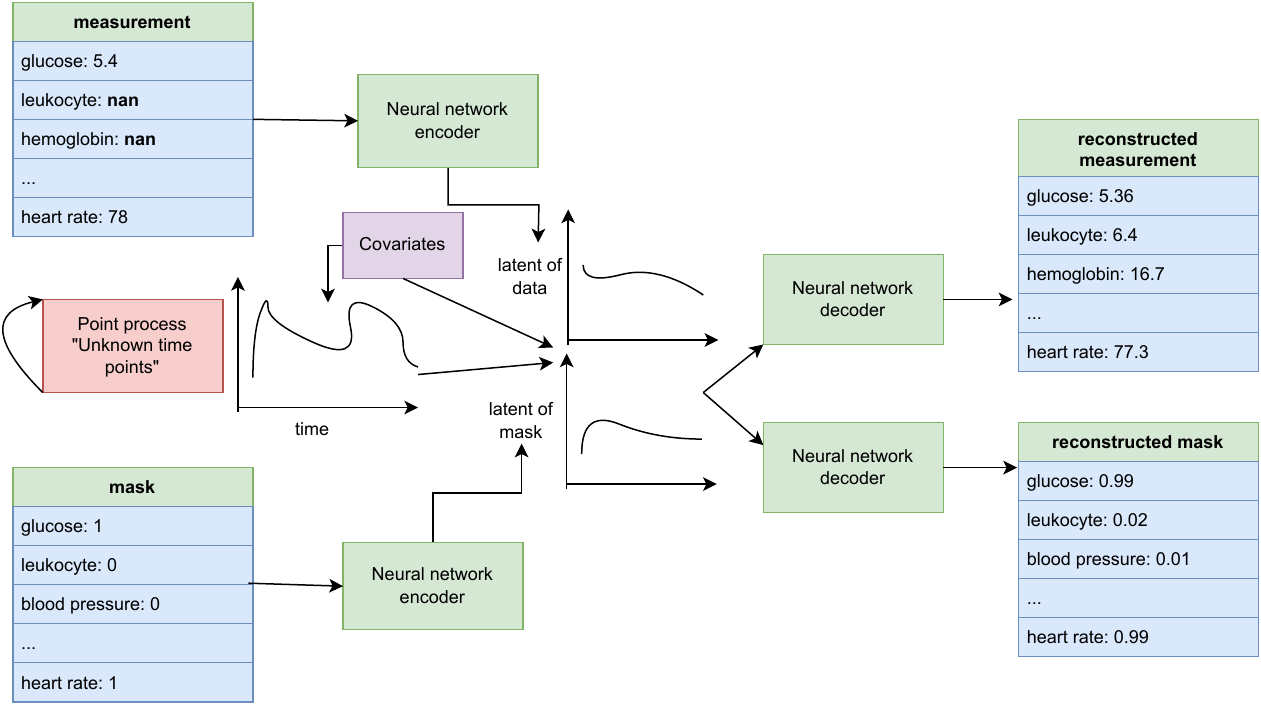}
    \caption{Detailed overview of our model}  
\label{fig:detailed_overview}
\end{figure*}

\begin{figure*}[!h]
    \centering
\includegraphics[width=0.8\textwidth]{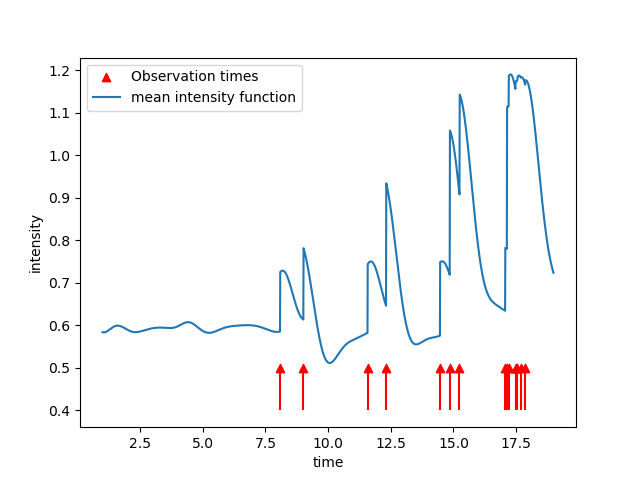}
    \caption{Inferred mean intensity function of the point process}  
\label{fig:mean_intensity}
\end{figure*}

\end{document}